\documentclass[sn-standardnature,iicol,a3paper]{sn-jnl}% Standard Nature Portfolio Reference Style
\jyear{2021}%

\theoremstyle{thmstyleone}%
\theoremstyle{thmstyletwo}%

\theoremstyle{thmstylethree}%

\usepackage{dsfont}
\usepackage{subcaption}

\begin{document}

%\larg

\title[Article Title]{Neuromorphic Computing with AER using Time-to-Event-Margin Propagation}

%%=============================================================%%
 %Prefix	-> \pfx{Dr}
 %GivenName	-> \fnm{Joergen W.}
 %Particle	-> \spfx{van der} -> surname prefix
 %FamilyName	-> \sur{Ploeg}
 %Suffix	-> \sfx{IV}
 %NatureName	-> \tanm{Poet Laureate} -> Title after name
 %Degrees	-> \dgr{MSc, PhD}
 \author[1]{ \fnm{Madhuvanthi Srivatsav R}} 
%\equalcont{\small These authors contributed equally to this work. }
%\author[1]{ \fnm{Lakshmi Annamalai}} 
%\equalcont{\small These authors contributed equally to this work.}
\author*[2]{ \fnm{ Shantanu Chakrabartty}} 
\author*[1]{ \fnm{ Chetan Singh Thakur}}
%\equalcont{All correspondence for this work should be addressed to {\it shantanu@wustl.edu} and {\it csthakur@iisc.ac.in} These authors contributed equally to this work.}
%%=============================================================%%

\affil[1]{\small \orgdiv{Department of Electronic Systems Engineering}, \orgname{Indian Institute of Science}, \orgaddress{ \country{India}}}

\affil[2]{\small \orgdiv{ Department of Electrical Engineering and Systems Engineering}, \orgname{Washington University},\state{ St. Louis}, \country{USA}}
\affil[*]{\small Contact Address: {\it shantanu@wustl.edu} and {\it csthakur@iisc.ac.in}}




\abstract{
Address-Event-Representation (AER) is a spike-routing protocol that allows the scaling of neuromorphic and spiking neural network (SNN) architectures to a size that is comparable to that of digital neural network architectures.  However, in conventional neuromorphic architectures, the AER protocol and, in general, any virtual interconnect plays only a
passive role in computation, i.e., only for routing spikes and events. In this paper, we show how causal
temporal primitives like delay, triggering, and sorting inherent in the AER protocol itself can be exploited
for scalable neuromorphic computing using our proposed technique called Time-to-Event Margin Propagation
(TEMP).

 The proposed TEMP-based AER architecture is fully asynchronous and relies on interconnect delays for memory and computing as opposed to conventional and local multiply-and-accumulate (MAC) operations. We show that the time-based encoding in the TEMP neural network produces a spatio-temporal representation that can encode a large number of discriminatory patterns.

As a proof-of-concept, we show that a trained TEMP-based
convolutional neural network (CNN) can demonstrate an accuracy greater than 99\% on the MNIST dataset. Overall, our work is a biologically inspired computing paradigm that brings forth a new dimension of research to the field of neuromorphic computing.
}

\maketitle

\section{Introduction}

\let\thefootnote\relax\footnote{All correspondence for this work should be addressed to {\it shantanu@wustl.edu} and {\it csthakur@iisc.ac.in}}

Address-Event-Representation (AER) is a popular event-based asynchronous protocol used commonly in the design of large-scale and re-configurable neuromorphic hardware \cite{boahen2000point, bamford2010large, rathi2022exploring}. AER uses packet-based switching and time-division-multiplexing to achieve brain-scale connectivity on 2-dimensional and 2.5-dimensional hardware platforms, which are limited by the number of physical interconnects and routing pathways. In literature, different variants of the AER communication protocol have been proposed to improve channel capacity \cite{bamford2010large} and system scalability by reducing memory requirements~\cite{rathi2022exploring, Jongkil}. However, in most previous implementations, AER and other interconnect mechanisms (virtual and physical) have only played a passive role, i.e., they only transmit signals. In such architectures, spike-routing latency is viewed as a nuisance or a source of system uncertainty. But neurobiology suggests otherwise. Recent research has shown that neuronal dendrites, which can be viewed as the neurobiological equivalent of interconnects, exhibit a range of linear and nonlinear mechanisms that allow them to implement elementary computations \cite{michael2005dendritic, izhikevich2006polychronization, sun2022axonal}. These findings have inspired spiking neural networks (SNNs) architectures using active interconnects or interconnects with computational capabilities \cite{galluppi2011representing, sun2022axonal}. Also, a recently proposed dendro-centric computing framework \cite{boahen2022dendrocentric} extends the concept of active interconnects further and proposes to encode information spatio-temporally in the pulse or spike sequences. These sequences can then be decoded using nano dendrites, and this modality can be used to address specific neurons using the sequences as addresses. From an energetic point of view, this kind of information processing has been proposed to scale linearly with the number of neurons, thus enabling energy-efficient AI applications \cite{boahen2022dendrocentric}. Another major argument for incorporating processing-in-interconnects through axonal or dendritic delays is the premise that both spatial and temporal encoding can produce 
different groups of neurons that fire in specific temporal sequences. Such networks with axonal delays have enormous memory capacity as they have more groups than neurons (due to combinatorial factors) and thus could exhibit a massive diversity in network responses \cite{izhikevich2006polychronization}. Further, by introducing re-configurable and trainable axonal delays, we can eliminate the need for synaptic circuits, thus ensuring compact hardware architectures.
\newline

\begin{figure*}
\centering
\includegraphics[width=\textwidth]{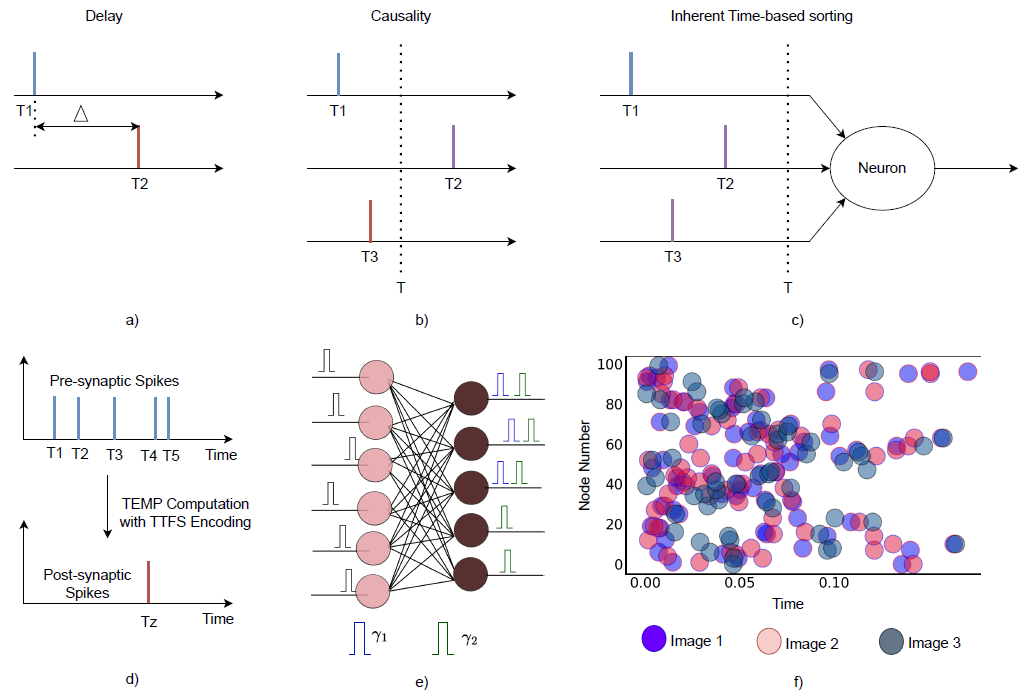}
\caption{\textbf{Proposed TEMP-based computing paradigm:} Virtual interconnects (AER) can only a) delay pulses (spikes) or b-c) determine causal relationships between pulses (time-based sorting).  d) The formulation of TEMP incorporates the TTFS encoding scheme, where the post-synaptic neuron processes the information from the pre-synaptic spikes and encodes its output in the precise spiking time of one spike.  e) The application-specific tunable $\gamma$ parameter controls the sparsity in encoding information. Here $\gamma_{1}$ $>$ $\gamma_{2}$, and it can be observed that as $\gamma$ is increased, sparsity is enhanced. f) The time-domain computations in TEMP bring forth spatio-temporal encoding of input patterns. Diverse encoding patterns can be observed for images of the same class in a network trained on the MNIST dataset. This diversity can be attributed to the fact that the information is encoded in the group of neurons that fire and the order in which they fire. }
\label{fig:proposed}
\end{figure*}

In this paper, we present a spike computing framework called time-to-event margin-propagation (TEMP) that exploits the computational primitives inherent in AER and other spike-routing or interconnect architectures. These are generally causal primitives like delay, trigger, and sorting operations, as shown in Fig.~\ref{fig:proposed}(a)-(c), which can be easily implemented using time-division-multiplexing and packet-switching networks. For instance, the triggering operation illustrated in Fig.~\ref{fig:proposed}(b) passes an input spike only if it arrives before a specific time instant denoted by T. Similarly, sorting Fig.~\ref{fig:proposed}(c) is naturally implemented because of the temporal ordering of spikes. Using TEMP, we show that these fundamental operations can be used to demonstrate non-linear classification abilities producing competitively comparable results to traditional multi-layer neural networks. Further, TEMP models the information in the precise timing of the spikes, with the help of TTFS (Time to First Spike) encoding  Fig.~\ref{fig:proposed}(d).  TTFS coding leads to a prominent reduction in inter-neuron spikes, thus curtailing the energy consumption in information transmission.  A hyper-parameter in the TEMP formulation controls the network's sparsity, latency, and accuracy, thus ensuring its adaptability to diverse applications. As highlighted in Fig.~\ref{fig:proposed}(e), by tuning the hyper-parameter $\gamma$, a TEMP network can control the number of output spikes/sparsity of a layer. This formulation can be used to realize a much richer M-of-N spike encoding \cite{galluppi2011representing} or K-based encoding strategies \cite{boahen2022dendrocentric}. Additionally, the asynchronous nature of TEMP allows the network to encode information using temporal dynamics that results in spatio-temporal encoding of features, which exhibits enormous memory capacity. This is illustrated in Fig.~\ref{fig:proposed}(f), where a TEMP network trained to discriminate digits exploits different spike-timing patterns involving different groups of neurons for images of the same digit. 

\section{Results}

\paragraph{Event-based Model for a TEMP Neuron}

At the core of TEMP is margin propagation which is a piece-wise-linear approximate computing technique introduced in \cite{ChakrabarttyCauwenberghs,MingShantanu} and extended in  \cite{NairChakrabartty,NairNath}. TEMP extends margin propagation into the time domain where a TEMP neuron generates a spike/event at time instant $t$ when the following condition is satisfied

\begin{equation}
\sum_j\vert{t-t_j}\vert_+=\gamma
\label{eq:R2}
\end{equation}

\begin{figure*}
\centering
\includegraphics[width=\textwidth]{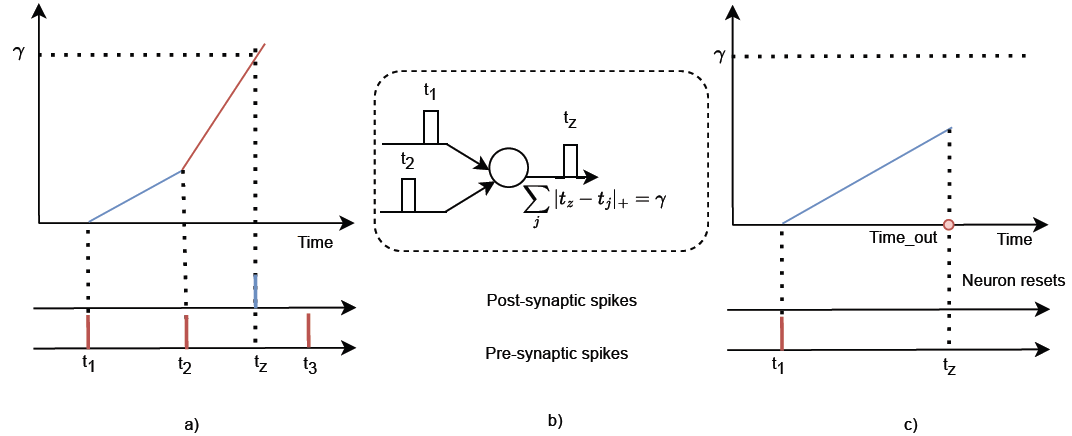}
\caption{\textbf{Computational intricacies of TEMP-based neurons:} a) As the post-synaptic neuron receives pre-synaptic spikes at time $t_1$ and $t_2$, its membrane potential starts rising with increasing slopes determined by the number of pre-synaptic neurons that it encounters. b) Network representation of TEMP-based neurons. c) Every TEMP neuron is associated with a time-out, after which it will no longer spike and will reset.}
\label{fig:temp}
\end{figure*}

Here $t_j$ denotes the arrival time of the $j^{th}$ pre-synaptic spike/event, $\gamma > 0$ denotes the firing threshold, and $[.]_+$ denotes a ReLU function. Fig.~\ref{fig:temp}(a,b) shows a possible mechanism for implementing equation~\ref{eq:R2}. An internal state variable (for example, a counter or a capacitor) stores the membrane potential, which is updated every instant a pre-synaptic spike/event occurs $t_1, t_2,...$. However, at the $j^{th}$ event, the state variable or the counter update rate is increased by $j$. Thus, as more events arrive, the state-variable increases at a faster rate. When the state variable reaches the threshold value $\gamma$, say at time instant $t_{z}$, the TEMP neuron emits a spike. The ReLU operation in equation~\ref{eq:R2} is naturally implemented due to time-causality - that is, any spikes that arrive after $t_{z}$ are ignored during the computation as shown in Fig.~\ref{fig:temp}(c). 
Also, every neuron is associated with a $time_{out}$ factor at which it will reset its counter or state variable. If the neuron's potential does not reach the threshold or $\gamma$ before the $time_{out}$, the neuron will no longer spike and will be reset.
Note that while there could be several techniques to implement TEMP on digital, analog, electronic, and non-electronic hardware, this paper focuses on the system architecture and not on specific implementation details. 

\paragraph{TEMP spiking neural network}

Like other SNN architectures, two TEMP neurons $i$ and $j$ can be connected to each other using a synaptic weight $w_{ij}$. However, unlike the conventional SNN formulations, the role of synaptic weights in the TEMP network is to delay the input spikes. Following a differential margin-propagation architecture proposed in \cite{NairNath}, \cite{NairChakrabartty} to approximate inner-products, a similar mapping is also applied to equation~\ref{eq:R2}. The output of an $i^{th}$ neuron in a TEMP network is two spikes/events denoted by their respective time of occurrence $t_i^+$ and $t_i^-$. These occurrences are computed according to the following:

\begin{align}
\sum_j\vert{t_i^+-(t_j^++w_{ij}^+)}\vert_++\vert{t_i^+-(t_j^-+w_{ij}^-)}\vert_+=\tau_m \nonumber \\
\sum_j\vert{t_i^--(t_j^++w_{ij}^-)}\vert_++\vert{t_i^--(t_j^-+w_{ij}^+)}\vert_+=\tau_m
\label{eq:R4}
\end{align}

Here, the synaptic weights are represented as differential quantities as $w_{ij} = w^+_{ij} - w^-_{ij}$, with $w^+_{ij},w^-_{ij} \ge 0$. Both the positive quantities $w^+_{ij},w^-_{ij}$ are time-delays which ensures that the equations~\ref{eq:R4} are causal. The occurrence times $t_i^+$ and $t_i^-$ are then processed according to a
differential ReLU operator, which is given by
\begin{equation}
(t_i^+, t_i^-) =\left\{ 
  \begin{array}{ c l }
    (t_i^+,t_i^-) & \textrm{if } t^+ \geq t^- \\
    (t_i^-,t_i^-) & \textrm{otherwise}
  \end{array}
\right.
\label{eq:R4a}
\end{equation}
There exists some equivalence/correspondence between the TEMP event-based model and the leaky-integrate-fire (LIF) neuronal network, which is described in ~\ref{sec:appendix1} (Appendix I).
\begin{figure*}
\centering
\includegraphics[width=\textwidth]{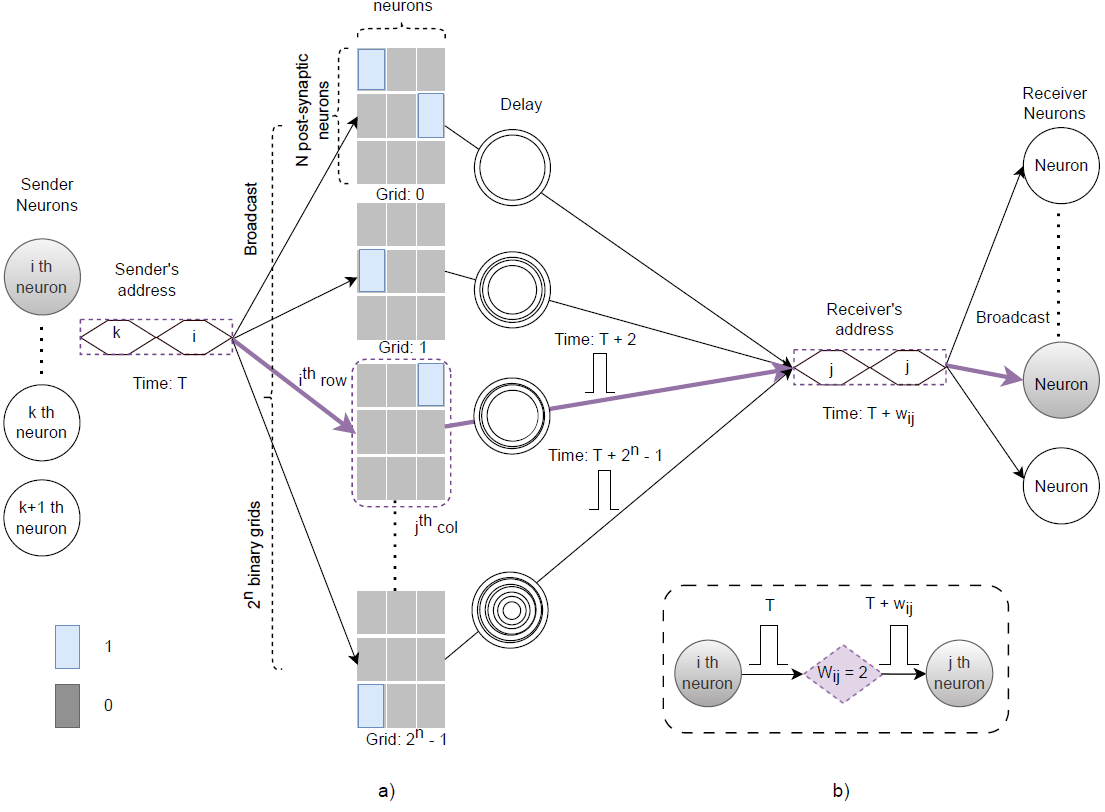}
\caption{\textbf{Realization of TEMP in AER:} a) As represented in the connectivity representation figure (b), the $i^{th}$ sender neuron with address i, sends out a spike, which needs to be transmitted to the $j^{th}$ receiver neuron with a delay of 2-time units. The sender's spike event is broadcasted to the $2^q$ (N x M) binary routing tables, which contain the connectivity information. In the second routing table (Grid:2), a 1 is present at the grid (i,j), which indicates a connection between the $i^{th}$sender neuron and the $j^{th}$ receiver neuron (0 indicates no connection). Accordingly, a spike event with the receiver's address is sent to the digital bus after a delay of 2-time units. b) Connectivity information representing the connectivity between the $i^{th}$ and $j^{th}$ neuron with an interconnect delay of 2-time units. } 
\label{fig:AER}
\end{figure*}

\paragraph{AER Realization of TEMP networks}

Here we describe a possible mechanism to implement TEMP networks using the AER protocol. We will assume access to the trained parameters $w_{ij}$, which is assumed to be quantized (or can assume only specific values). It has been observed that post-training quantized weights (at a precision equal to or greater than 8 bits) provide the same level of recognition accuracy as a network with full-precision weights. Therefore, for q-bit quantization, each of the weights can assume $2^q$ possible values. For the proposed implementation, we will instantiate $2^q$ routing tables of size $N \times M$, where $N$ is the number of post-synaptic/receiver neurons and $M$ is the number of pre-synaptic/sender neurons. This is shown in Fig.~\ref{fig:AER}(a), where an entry in each of the routing tables is a $1$ or $0$ entry indicating if a sender neuron emits a spike, the event is routed to the destination neuron after a fixed delay. Note that the delay corresponding to each routing table is fixed, and all routing tables share a common output bus/interconnect. The AER protocol is then used by the destination (or post-synaptic) TEMP neurons to receive the event, which then process information according to equation~\ref{eq:R2}. The fixed delay in Fig.~\ref{fig:AER}(a) could be implemented using physical interconnects or using time-outs. Specific implementation details will be a topic for another paper.

\paragraph{Network topology constructed with TEMP}

To demonstrate the advantages achieved with the proposed TEMP framework when applied to machine learning tasks, a population of TEMP neurons are connected with each other in a feed-forward fashion. The results are based on the implementation of TEMP as given by Eq. \ref{eq:R4} in the spiking network using a standard deep learning framework.

Spatio-temporal input stimuli are interfaced to the network through a population of neurons which we call sensory neurons. The sensory layer is projected onto the subsequent layer through learnable conduction delays. Multitude sets of neurons in this layer respond to unique sequences in the stimulus, resulting in high dimensional spatio-temporal firing patterns. To verify the representational capability, the generated spatio-temporal patterns are projected onto the neurons of the recognition layer. The classifier TEMP neuron that fires differential spikes with minimal delay between them is declared the winner class.

\paragraph{Non-linear Classification using TEMP}

\subparagraph{\textbf{MNIST classification task}}

To investigate the credibility of the proposed TEMP network in terms of generalization capability, we applied it to the prevalent MNIST classification task. The network architectures used consist of TEMP-based fully-connected layers and convolution layers. Pixel intensities were translated to differential spike trains.

For the architecture implementing a $784\rightarrow100\rightarrow10$ network, the dense layer was trained in an end-to-end supervised fashion with a batch size of $32$ using the ADAM optimizer with an initial learning rate of $0.001$ to minimize the standardized categorical entropy loss. Training converges in $30$ epochs, reaching a best test accuracy of $97.7\%$.

For the architecture implementing a (28x28x1)$\rightarrow$(3x3x6)$\rightarrow$15$\rightarrow$10 convolution network,  training was done with mini-batches of size $32$ for $30$ epochs with Adam optimizer and an initial learning rate of $0.00005$. Batch normalization was implemented at the output layer.  This network achieved a best accuracy of $97.71\%$ when projected onto the classifier layer. By introducing batch normalization between successive layers as well, we were able to achieve $99.1 \%$ accuracy with two convolution layers (3x3 kernels with $16$ and $32$ channels ) followed by a hidden layer with $500$ and $10$ nodes. Each convolution layer was followed by a max pool layer. The 2-convolution layer network was trained with an ADAM optimizer with an initial learning rate of 0.001 and a batch size of 16. A time-based learning rate decay scheduler was used for all the training runs. 

\subparagraph{\textbf{Comparison with state-of-the-art}}

For comparison, we consider the results obtained with other spiking architectures. Table. \ref{table:sta} displays accuracy at par with that of state-of-the-art spiking architectures. Note that rate-based approaches display better accuracy but with sacrifice in latency, sparsity, and energy efficiency when compared to TTFS-based approaches. Note that we were able to achieve good accuracy with $100$ hidden nodes in a fully connected spiking network and $6$ channels with $15$ hidden nodes in a convolutional spiking network.

\begin{table}
\centering
\begin{tabular}{c c | c c}
\hline
\textbf{Method} & \textbf{Accuracy} & \textbf{Method} & \textbf{Accuracy}\\ \hline
\multicolumn{4}{c}{\textbf{Rate coding}} \\ \hline
\cite{Srinivasan} & 0.91 & \cite{Kheradpisheh} & 0.98 \\
\cite{ChuaZhang} & 0.984 & \cite{Stromatias} & 0.95 \\ \hline
\multicolumn{4}{c}{\textbf{TTFS coding}} \\ \hline
\cite{Mostafa} & 0.97 & \cite{ZhangZhou} & 0.99 \\
\cite{ZhouShibo} & 0.99 & \cite{Kheradpisheh} & 0.97 \\
\cite{Comsa} & 0.979 & \cite{Goltz} & 0.97 \\ 
TEMP$^1$ & \textbf{0.977} & TEMP$^2$ & \textbf{0.991/0.999} \\ \hline
\end{tabular}
\caption{Comparison of proposed TEMP with state-of-the-art spiking architectures on MNIST classification task. TEMP$^1$ and TEMP$^2$ are spiking networks with dense and convolutional layers, respectively. TEMP$^2$ (test/train) is the result obtained with normalized spatio-temporal patterns. Rate-based approaches (top $2$ rows) lack the benefits such as sparsity, energy efficiency, etc., which are inherent to TTFS-based approaches.}
\label{table:sta}
\end{table}

\paragraph{Analysis of Spatio-Temporal encoding in TEMP Networks}

\begin{figure*}
\centering
\begin{subfigure}{\textwidth}
\centering
\includegraphics[width=0.22\textwidth]{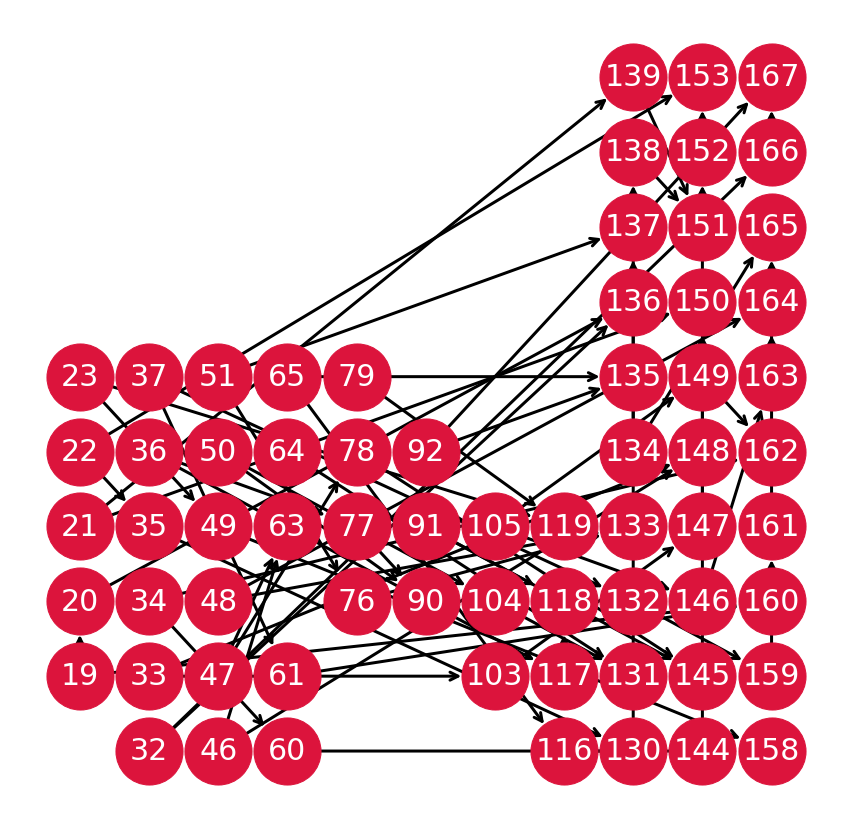}
\includegraphics[width=0.22\textwidth]{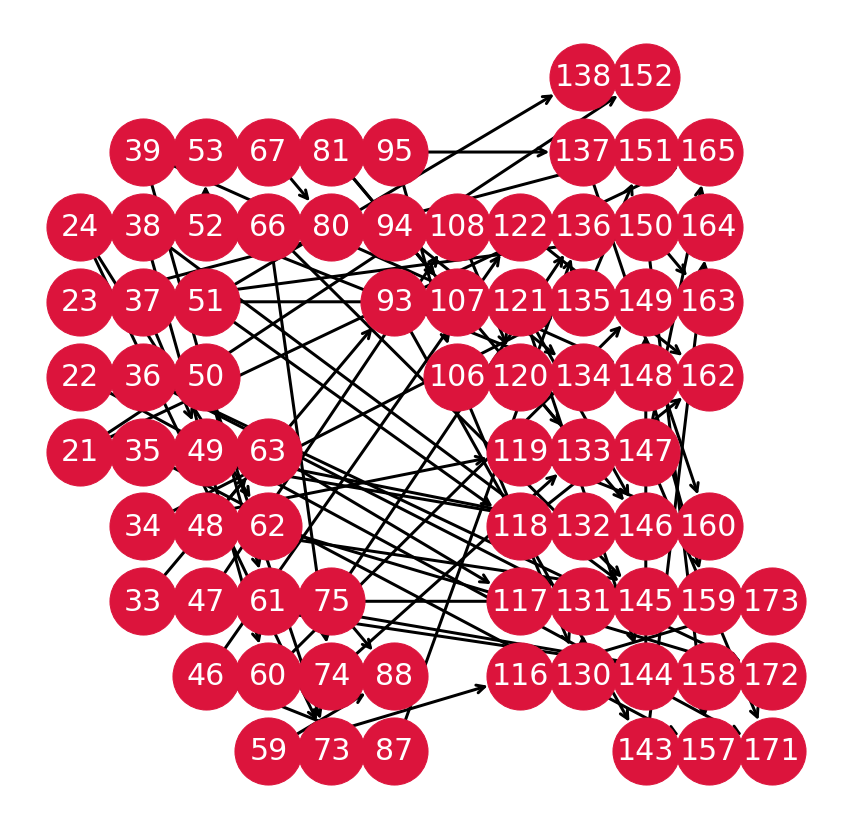}
\includegraphics[width=0.22\textwidth]{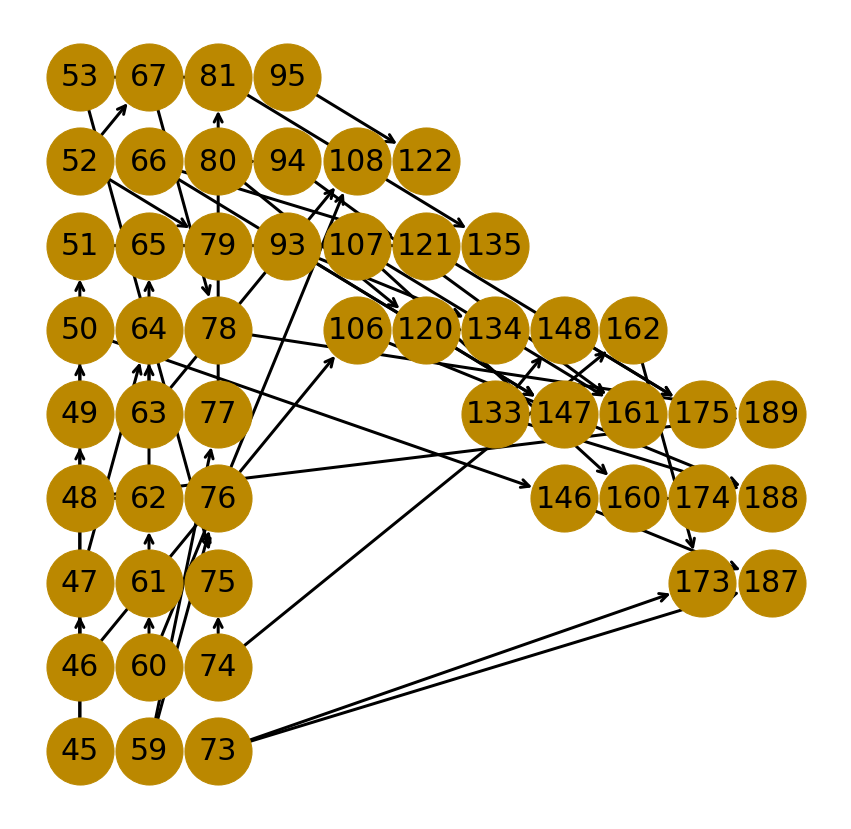}
\includegraphics[width=0.22\textwidth]{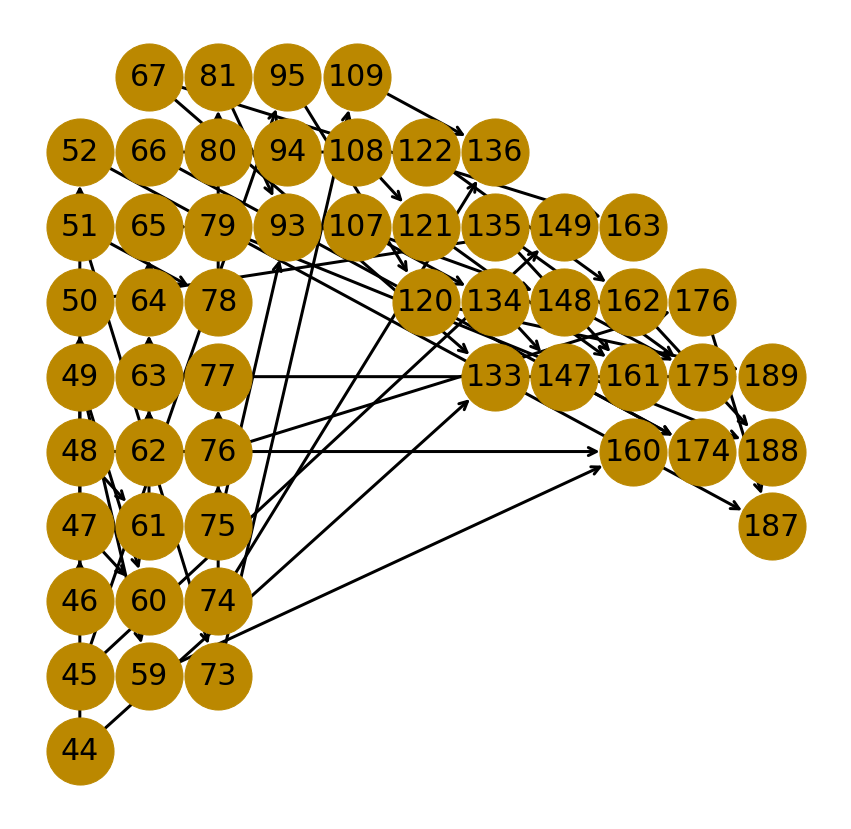}
\includegraphics[width=0.22\textwidth]{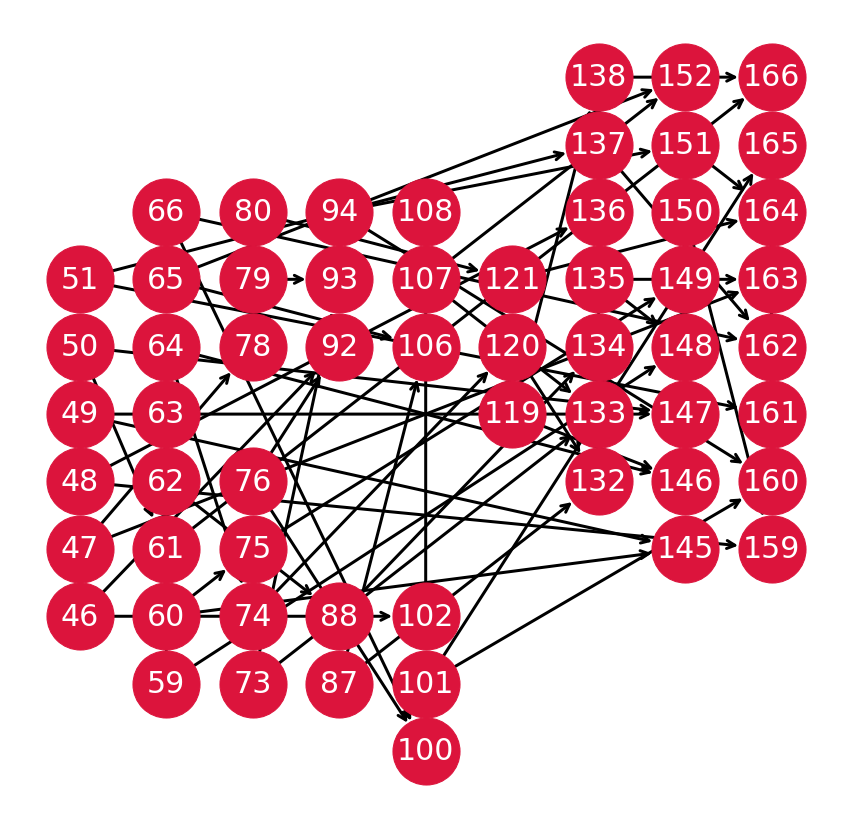}
\includegraphics[width=0.22\textwidth]{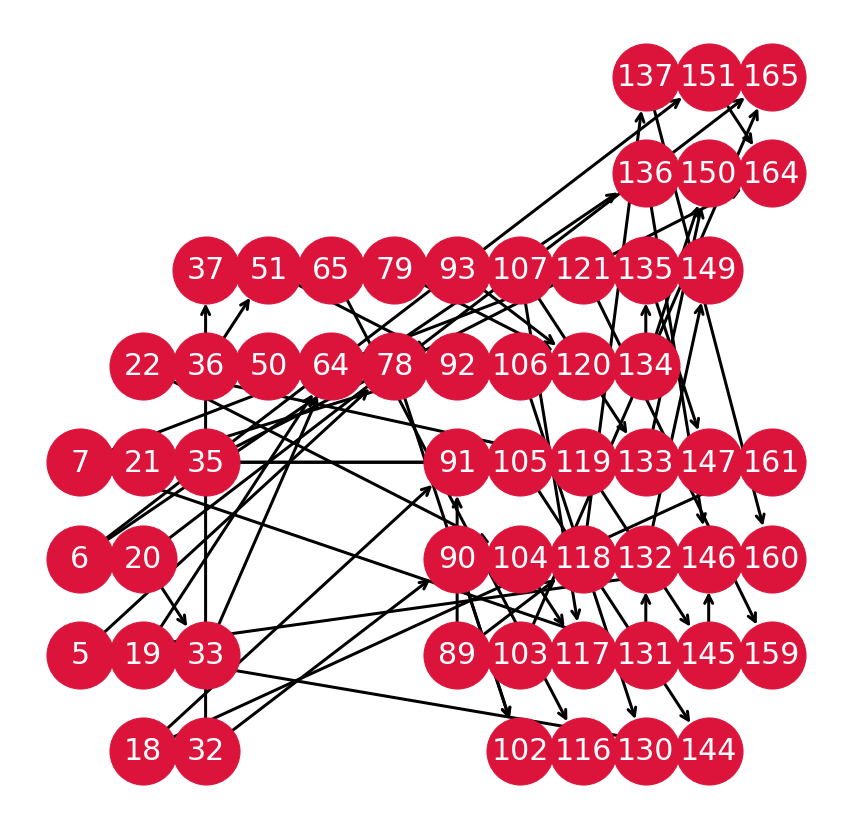}
\includegraphics[width=0.22\textwidth]{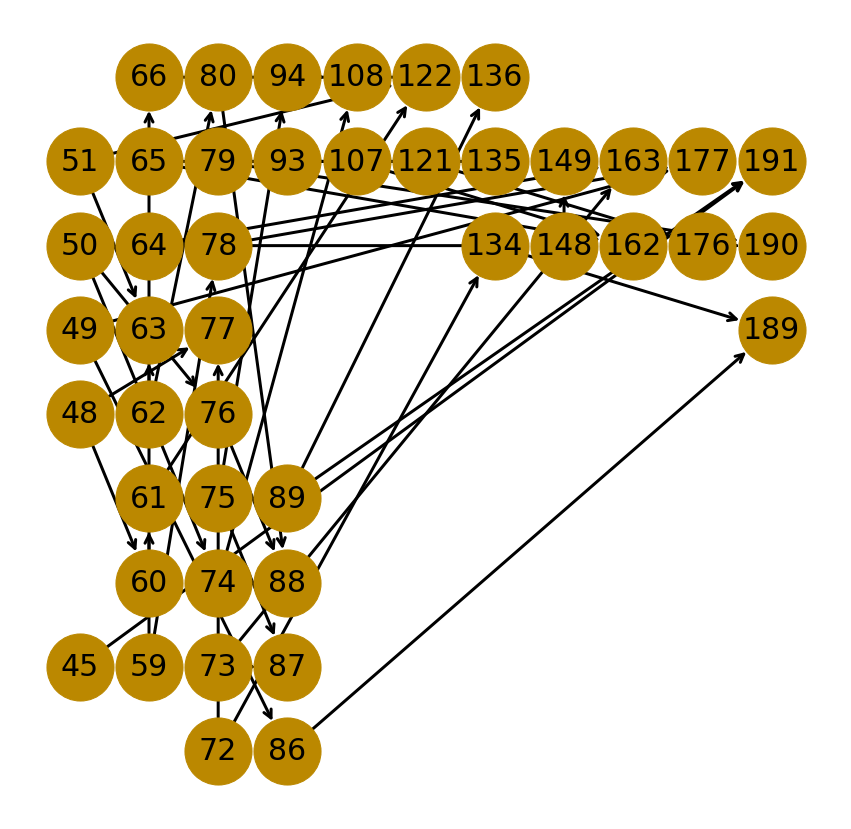}
\includegraphics[width=0.22\textwidth]{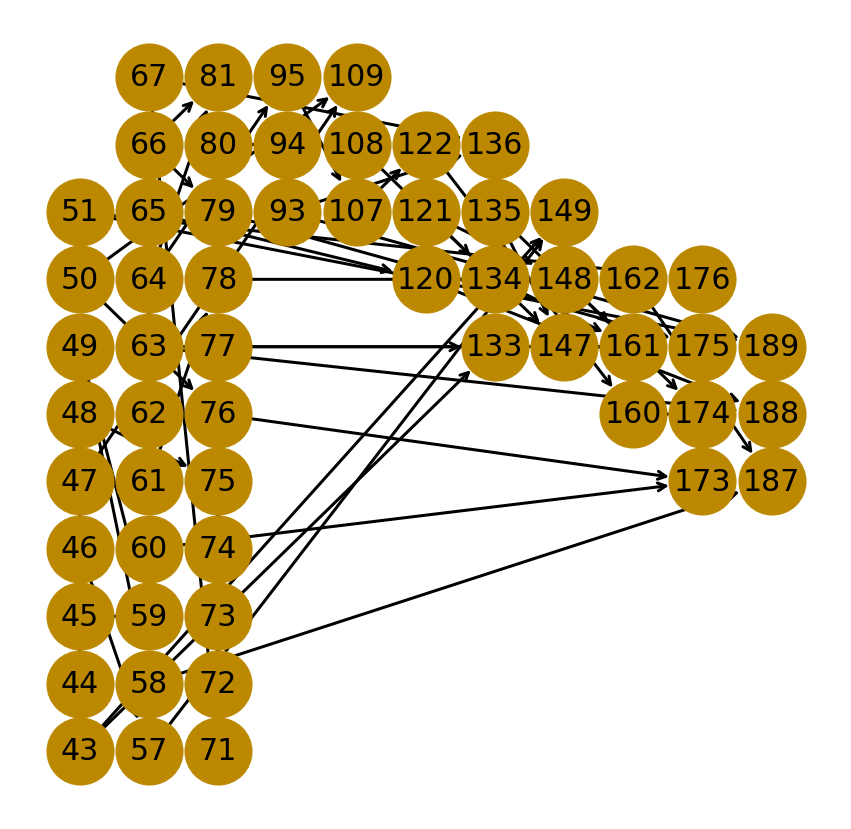}
\newline
\text{\small a)}
\end{subfigure}

\begin{subfigure}{\textwidth}
\centering

\includegraphics[width=0.32\textwidth]{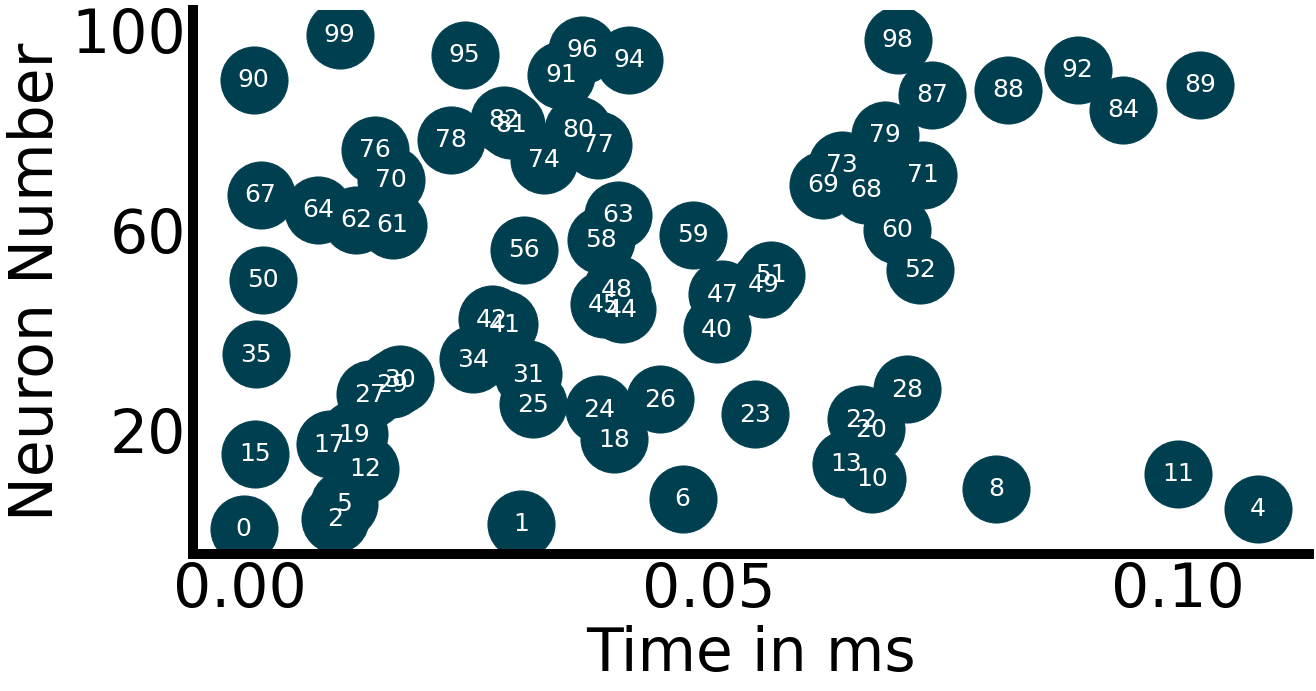}
\includegraphics[width=0.32\textwidth]{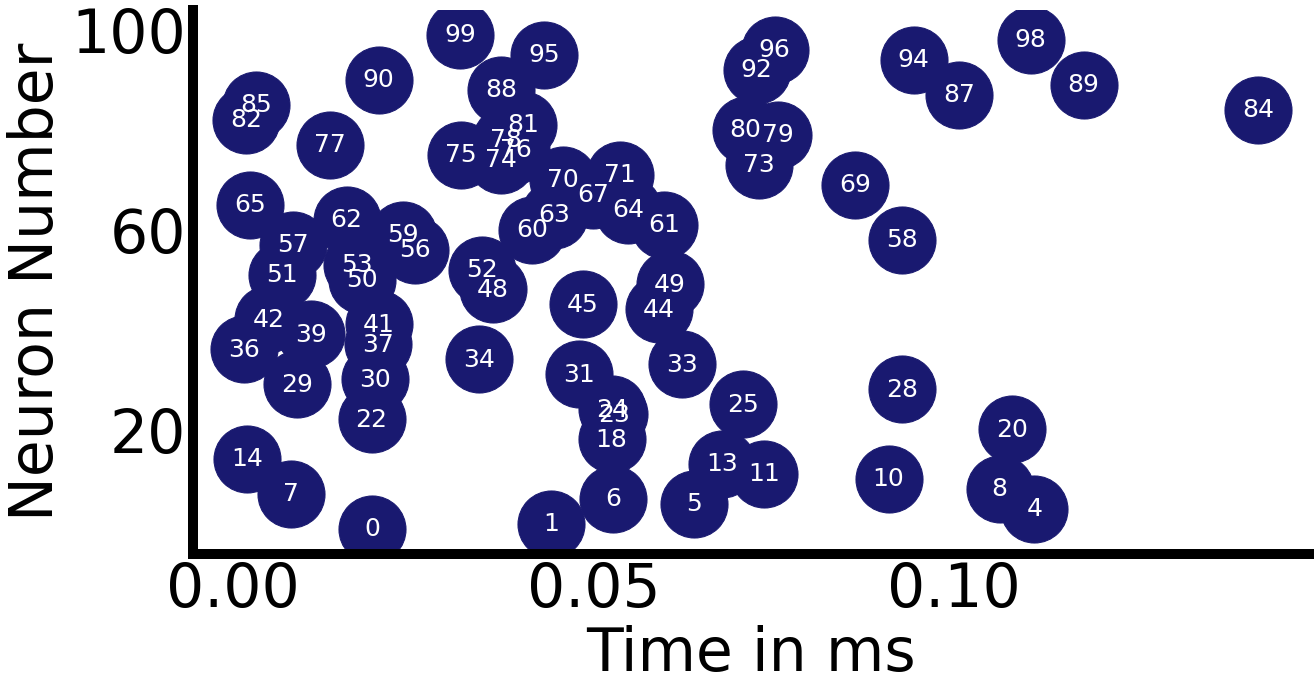}
\includegraphics[width=0.32\textwidth]{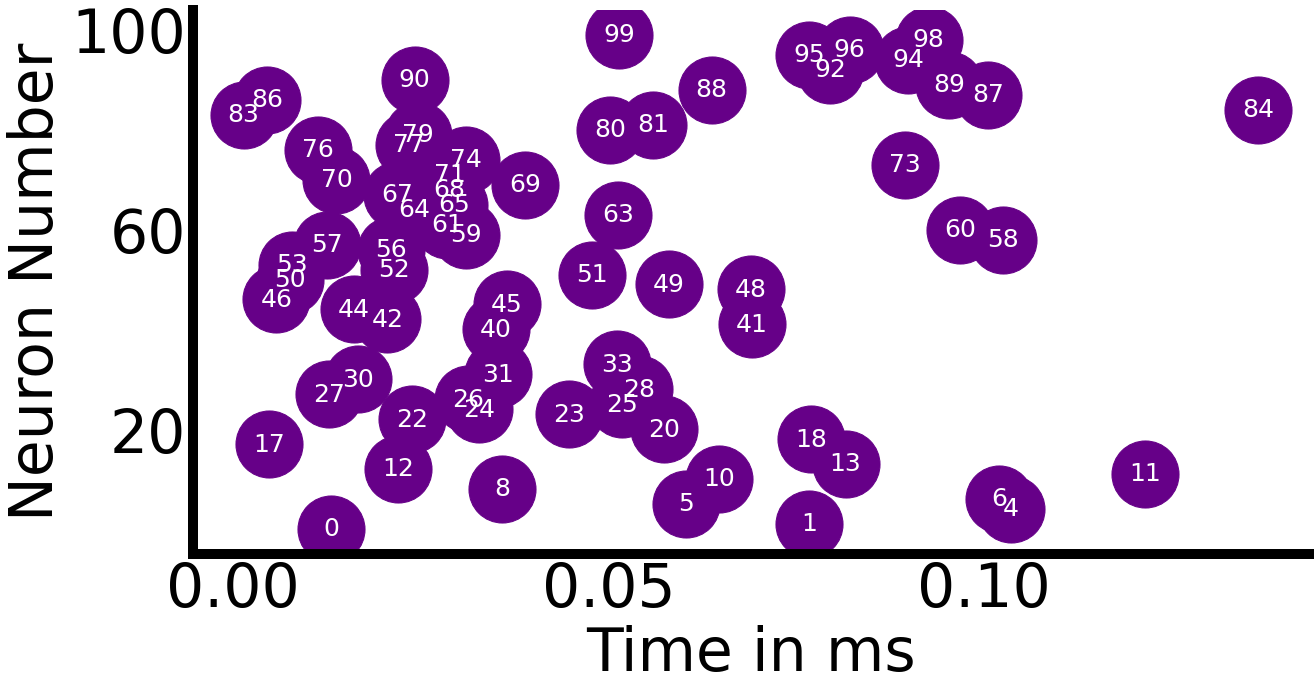}
\centering
\newline
\text{\small b)}
\end{subfigure}
\caption{\textbf{Combinatorial representational capability of spatio-temporal patterns.} (a) Spatio-temporal patterns emerged from the TEMP-based convolution layer in response to samples from input stimuli representing classes $2$ and $7$. This shows the intra-class similarity and inter-class variability learned by the patterns. Despite the intra-class similarity, unique spatio-temporal firing patterns can be observed for each stimulus, thus bringing out the combinatorial representational capability of TEMP.  (b) Spike raster plot of the learned spatio-temporal patterns. It could be seen that patterns share neurons, but the neuronal firing order differs across the patterns. This adds to the combinatorial representational capability of TEMP-based networks.}
\label{fig:combinePoly}
\end{figure*}

By virtue of axonal delays, TEMP-based networks exhibit rich spatio-temporal encoding, which is well-known for their enhanced combinatorial representational capabilities. As the network becomes structured with learning, certain stimuli-specific patterns emerge as portrayed in Fig. \ref{fig:combinePoly}, validating the combinatorial representational capability of these encoding patterns.

As noticed in Fig. \ref{fig:combinePoly}a, though there exists a similarity between patterns that belong to the same class, patterns that emerged in response to different stimuli do exhibit variance. 

Fig. \ref{fig:combinePoly}b shows the spike-raster plot (Time of firing vs. Neuron Number) of a dense layer. This experiment is an indication of the fact that there will be no ambiguity even if a set of neurons is shared across multiple spatio-temporal patterns. This is owing to the fact that a single neuron can fire with different patterns at different times, and these patterns are not only defined by their constituent neurons but also by their precise firing time.

\begin{figure*}
\centering
\includegraphics[width=\textwidth]{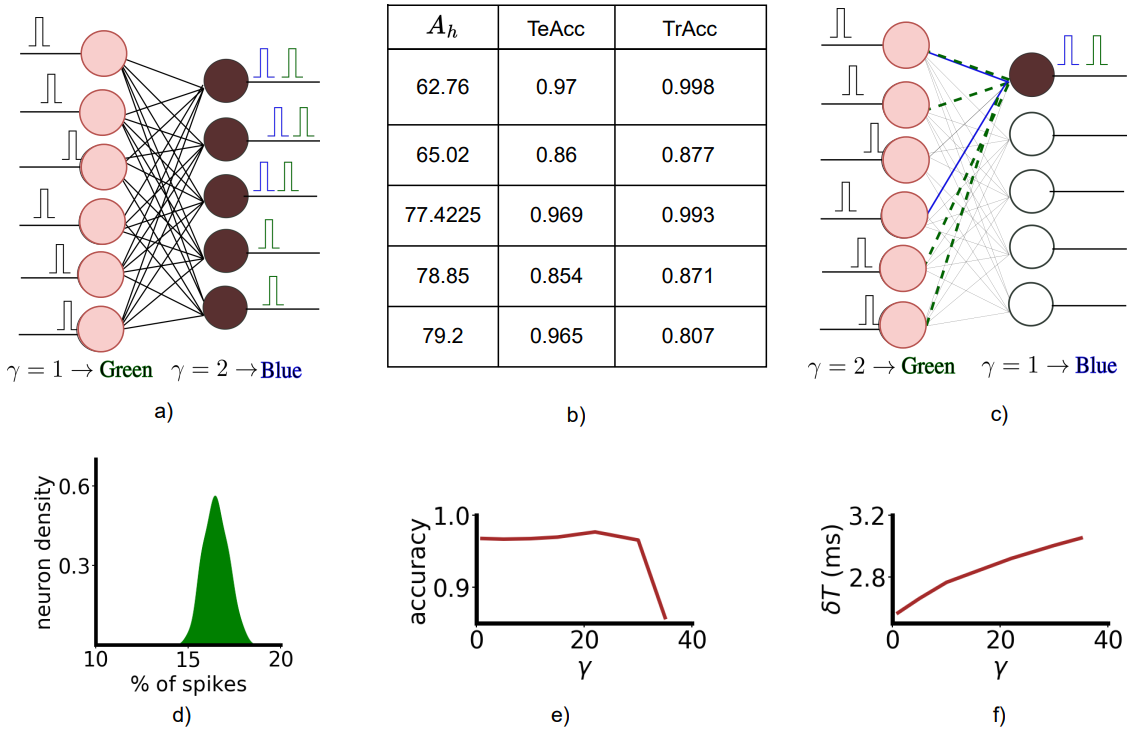}
\caption{\textbf{Effect of hyperparameter $\gamma$ on computation} (a,b,c) Sparse coding: (a) The hyperparameter $\gamma$ determines the number of nodes that fire, which is a crucial factor in achieving desirable sparsity. (b) Sparse coding vs. accuracy analysis. $A_h$ is the percentage of active nodes, which is controlled by the hyperparameter $\gamma$. TrAcc and TeAcc are the train and test accuracies at various sparsity coefficients. (c,d) Sparse causal spikes: (c) Based on $\gamma$, the causal set of pre-synaptic spikes varies, which determines the number of pre-synaptic spikes involved in the firing of the post-synaptic neuron. The delay plasticity ensures that most informative pre-synaptic neurons get to deliver their spikes early to post-synaptic neurons. This results in minimal computation. (d) This gives the plot of the distribution of the number of causal pre-synaptic spikes required by the TEMP neurons to fire. The total number of pre-synaptic spikes was $784$. It could be seen that, on average, the hidden layer TEMP neurons require only $\sim16\%$ of total pre-synaptic spikes to get triggered. This proves that delay plasticity results in the early arrival of most informative spikes, hence reducing the computations to as minimal as possible. (e,f) Tunable latency vs. accuracy: Analysis of latency and accuracy of TEMP network trained over different values of $\gamma : [1\ldots{35}]$. As $\gamma$ increases, latency increases as expected. However, classification accuracy reaches its maximum for a particular value of $\gamma$ as the latter is the critical factor in tuning the non-linearity induced by the TEMP. Thus, $\gamma$ acts as a hyperparameter for the latency vs. accuracy tradeoff.} 
\label{fig:scalable}
\end{figure*}

%The signaling energy cost is given as $\#$ of active node ($A_h$) $\times{ac}$ \cite{Guanzheng}, where $A_h$ is the sparsity-induced active nodes in hidden layer with $100$ nodes and $ac$ is the signaling cost. Considering the $ac$ cost of CMOS neuron ($12.5pJ$) given in \cite{Gangotree}, the analysis is provided in Table. \ref{fig:scalable}c across varying classification accuracies. 
\subparagraph{\textbf{The Effect of the Hyper-parameter $\gamma$ on Computation}}
\subparagraph{\textbf{Sparsity in causal spikes}}
The majority of energy is consumed by signaling, which could be regulated by sparse coding, where $M$ out of total $N$ neurons ($M\leq{N}$) are active. Sparse coding is an energy-saving neural coding along the lines of neural signal transmission theory and energy utilization rate theory. 

The formulation of TEMP neuron includes a hyper-parameter named $\gamma$, which regulates the number of post-synaptic neurons that fires (Fig. \ref{fig:scalable}a). Thus, TEMP exhibits an adaptive sparse coding strategy, which enables it to generate sparse $M$ of $N$ spike codes.

The proposed TEMP has adaptive delay plasticity, thus introducing a scheme where the order of arrival of pre-synaptic spikes at a post-synaptic neuron is governed by their information content. In this study, we highlight the effect of delay plasticity and $\gamma$ on the processing ability of TEMP (Fig. \ref{fig:scalable}c).

Fig. \ref{fig:scalable}d displays the distribution of pre-synaptic spikes required to fire each of the hidden layer neurons. It could be verified that an average neuron fires with $16 \%$ of total pre-synaptic spikes, thereby validating that the spikes representing critical information reach the neuron early and thus achieve the desired result with as few spikes as possible. 

In Appendix~\ref{sec:appendix2}, we have provided additional experiments on a simulated dataset to substantiate the mentioned claim.

\subparagraph{\textbf{Tunable latency (TT scaling)}}

The hyperparameter $\gamma$ plays a notable role in tuning the tradeoff between latency and accuracy. We could achieve a notable reduction in latency ($\gamma$) with a slight degradation in performance as provided in Fig. \ref{fig:scalable}(e,f). With an increase in latency ($\gamma$), the recognition capability of TEMP improves. However, as $\gamma$ is increased further, we notice the drop in accuracy, owing to the fact that the non-linearity exhibited by the TEMP neuron is causality-induced ($\gamma$ dependant) non-linearity. An increase in $\gamma$ is limited by the fact that the causality should induce sufficient non-linearity to generate separable patterns. However, there is an optimal value of $\gamma$, where we can obtain both competitive accuracies as well as reasonable latency.

\section{Discussion}

We have proposed a time-based spike computing paradigm TEMP.  This builds on a principle known as Margin Propagation (MP), which has been introduced to approximate log-sum-exp as a piecewise linear function. We remark that TEMP involves only primitive operations such as time-based addition, subtraction, threshold operation, etc. 

TEMP is built on delay plasticity, which contributes towards a unique implementation of the popular AER protocol. By modeling synaptic strength as interconnect delay, TEMP reduces the demands on neuromorphic hardware by completely eliminating synaptic circuits, thus favoring the construction of highly reconfigurable large-scale neuromorphic spiking architectures.

Gradient based-learning approaches have remained incompatible \cite{Shulz} \cite{Morrison} \cite{Diehl} \cite{Falez} with spiking models until recently \cite{Mostafa} \cite{Kheradpisheh}.  By incorporating temporal coding into a differentiable non-leaky TEMP formulation, we showed that we were able to define a continuously differentiable expression between input and output spike times. This enabled exact error backpropagation through a network of neurons, unlike the conventional spiking domain where direct learning is still an open research problem \cite{Diehl} \cite{Rueckauer} \cite{Deng} \cite{Ding} \cite{Shrestha} \cite{Shulz} \cite{Morrison} \cite{DiehlCook} \cite{Falez}. 

We showed that the network constructed with the proposed spike computing model could solve non-linear classification tasks with accuracy comparable to that of the state-of-the-art. The property of learnable delay, inherent to TEMP, has led to the emergence of spatio-temporal patterns in the network. The combinatorial representational capability of these patterns has been demonstrated for different classes of stimuli.

  Further, through experiments, the benefits of the tunable hyperparameter $\gamma$ inherent to TEMP have been demonstrated. $\gamma$ can be used to control the latency, accuracy, and sparsity in TEMP-based networks.
This application-specific tunable property of $\gamma$ enables
widespread application of TEMP from ultra-fast
differential sensing systems to highly accurate
visual recognition systems.

\appendix
\section{Appendix I}
\label{sec:appendix1}

\paragraph{TEMP Formulation}

Similar to \cite{MingShantanu}, the exponential function $f=e^{-(t-t_i)}$ can be approximated using Piecewise Linear (PWL) approximation as follows:

Differentiating $f$ with respect to $t$, it becomes

\begin{equation}
\frac{df}{dt}=-e^{-(t-t_i)}
\label{eq:4}
\end{equation}

Approximating $-e^{-(t-t_i)}$ for $t \geq {t_i}$ with the Heaviside step function, $-\theta(t-t_i)$, we get the approximated differentiation as 

\begin{equation}
\frac{df_a}{dt} = -\theta(t-t_i)
\label{eq:5}
\end{equation} 

Integrating the above equation we get,

\begin{eqnarray}
f_a &=& c-(t-t_i)\theta(t-t_i) \nonumber \\
&=&c-\vert{t-t_i}\vert_+
\label{eq:6}
\end{eqnarray}

Where $c$ is the integration constant. Hence $f_a$ turns out to be the approximation for $e^{-(t-t_i)}$.

\paragraph{Relation of TEMP with non-Leaky Integrate and Fire (n-LIF) Networks } 
In this section, we mathematically show the connection between the TEMP formulation and n-LIF SNNs. The differential equation governing the dynamics of an n-LIF neuron~\cite{goltz2019training, mostafa2017supervised} is given by

\begin{equation}
C_m\frac{du}{dt}=I(t)
\label{eq:7}
\end{equation}

Here, $C_{m}$ is the membrane capacitance, $u$ is the membrane potential and $I(t)$ is the current injected into the neuron. 
\newline
When pre-synaptic neurons $i$ emit a spike $\delta(t)$, they pass through their respective synaptic connections, where their strength gets modified by synaptic strengths $w_{i}$ and convolved with the following exponential synaptic kernel function,

\begin{equation}
\kappa(t)=\theta(t)e^{-\frac{t}{\tau_s}}
\label{eq:8}
\end{equation}

Here $\tau_s$ is the synaptic time constant and $\theta(t)$ is the Heaviside step function.
\newline
Inputs are received in the form of spikes that induce a synaptic current. The full synaptic
current is given by a weighted sum over the synapses from pre-synaptic neurons (denoted by index i) to the post-synaptic neuron with the respective weight $w_{i}$.
Accordingly, the total pre-synaptic current amounts to $I_{syn}(t)$ =  $\sum_i{w_{i}\kappa(t-t_{i})}$. 
Assuming a 0 initial condition on the membrane potential, the response of the n-LIF can be expressed as \cite{mostafa2017supervised}:

\begin{equation}
u(t){\frac{C_{m}}{\tau_{s}}=\sum_i{w_{i}\theta(t-t_{i})(1 - e^{-\frac{t - t_i}{\tau_s}})}}
\label{eq:9}
\end{equation}

Approximating Eq. \ref{eq:9} with the approximation in Eq. \ref{eq:6}, we get
\begin{equation}
u(t){\frac{C_{m}}{\tau_{s}}}=\sum_i{w_{i}\theta(t-t_{i})(1 - c + \vert{t-t_i}\vert_+})
\label{eq:10}
\end{equation}
For simpler computation, consider $w_{i} = 1$ for all i, and $c=0$.
At the threshold voltage $v_{th}$, attained at time $t_{z}$, the neuron spikes, and Eq \ref{eq:10} simplifies to 
\begin{equation}
v_{th}{\frac{C_{m}}{\tau_{s}}}=\sum_i{(1 + \vert{t-t_i}\vert_+})
\label{eq:11}
\end{equation}

\paragraph{Relation of TEMP with Leaky Integrate and Fire Networks } 
In this section, we mathematically show the connection between the TEMP formulation and the Leaky Integrate and Fire (LIF) SNNs. The differential equation governing the dynamics of an LIF neuron~\cite{rathi2020enabling, goltz2019training} is given by 

\begin{equation}
C_m\frac{du}{dt}=\frac{1}{R}[u_{rest}-u]+I(t)
\label{eq:12}
\end{equation}

Here, $C_m$ is the membrane capacitance, $u$ is the membrane potential, $R$ is the membrane resistance, $u_{rest}$ is the resting potential, $I(t)$ is the current flowing into the neuron.

The total pre-synaptic current can be denoted as $I_{syn}(t)$ =  $\sum_i{w_{i}\kappa(t-t_{i})}$, which represents the weighted sum over the synapses from pre-synaptic neurons (denoted by index i) with weights $w_{i}$. Assuming a 0 initial condition on the membrane potential, the response of the LIF neuron for an exponential synapse kernel (Eq \ref{eq:8}) can be expressed as \cite{goltz2019training}:

\begin{equation}
u(t)={\frac{1}{C_m}} {\frac{\tau_m \tau_s}{\tau_m - \tau_s}} \sum_i{w_{i}\theta(t-t_{i})(e^{-\frac{t - t_i}{\tau_m}} - e^{-\frac{t - t_i}{\tau_s}} )}
\label{eq:13}
\end{equation}

On approximating, the exponential terms in Eq \ref{eq:13} using Eq \ref{eq:6}, we get
\begin{equation}
u(t)={\frac{K_1}{C_m}} \sum_i{w_{i}\theta(t-t_{i})(c1 - \frac{\vert{t-t_i}\vert_+}{\tau_m} + c2 + \frac{\vert{t-t_i}\vert_+ }{\tau_s})}
\label{eq:14}
\end{equation}

Here $K1 =  {\frac{\tau_m \tau_s}{\tau_m - \tau_s}}$. 
\newline
Assuming the constants c1 and c2 are 0, and $w_i$ is 1 for all i, for simpler computation, Eq \ref{eq:14} can be expressed as

\begin{eqnarray}
u(t)={\frac{K_1}{C_m}} \sum_i{\vert{t-t_i}\vert_+ (\frac{1}{\tau_s} - \frac{1}{\tau_m} )} \nonumber \\
={\frac{1}{C_m}} \sum_i{\vert{t-t_i}\vert_+ }
\label{eq:15}
\end{eqnarray}

At the threshold voltage $v_{th}$, attained at the time
$t_{z}$ when the neuron spikes, Eq \ref{eq:15} simplifies to
\begin{equation}
v_{th} C_m= \sum_i{\vert{t_z-t_i}\vert_+}
\label{eq:16}
\end{equation}

\paragraph{LIF Neuron dynamics with a Dirac-Delta Synaptic Kernel}
 For $u_{rest}=0$ and an input current $I(t)=\delta(t)$, the impulse response of a LIF neuron  (from Eq \ref{eq:12}) becomes,

\begin{equation}
u(t)=\frac{1}{C_m}\theta(t)e^{-\frac{t}{\tau_m}}
\label{eq:17}
\end{equation}

Where $\tau_m=R_mC_m$ is the membrane time constant and $\theta(t)$ is the Heaviside step function. Assuming the synaptic impulse response to be a Dirac delta function and $w_{i}$ to be small, the response of the LIF neuron to a weighted sum of pre-synaptic spikes (from Eq. \ref{eq:13}) can be approximated as

\begin{equation}
u(t)=\frac{1}{C_m}{\sum_j\theta(t-t_j)\left[e^{-\frac{(t-t_j-w_j)}{\tau_m}}\right]}
\label{eq:18}
\end{equation}

Substituting the exponential term with its approximation $f_a$, we get,

\begin{equation}
u(t)=\frac{1}{C_m}{\sum_j\theta(t-t_j)\left[c-\vert{t-t_j-w_j}\vert_+\right]} 
\label{eq:19}
\end{equation}

When the neuron fires at $u(t)= v_{th}$, 

\begin{equation}
v_{th}=\frac{1}{C_m}\sum_j\left[c-\vert{t_{z}-t_j-w_j}\vert_+\right] 
\label{eq:20}
\end{equation}

Eq. \ref{eq:20} can be simplified as,

\begin{equation}
\sum_j\vert{t_{z}-t_j-w_j}\vert_+=\gamma =\sum_jc-{C_m}v_{th}
\label{eq:21}
\end{equation}

%\begin{eqnarray}
%&\sum_j\vert{t-t_j-w_j}\vert_+=\gamma \nonumber \\
%&Where, \gamma=\sum_jc-{C_m}\vartheta
%\label{eq:14}
%\end{eqnarray}

Note in Eq. \ref{eq:21} the synaptic connectivity between neurons has become axonal delay.

\paragraph{Inter-neuron connectivity of $\mathds{T}$\normalfont\textsc{emp}}

$\mathds{T}$\normalfont\textsc{emp} with synaptic connectivity is defined by the following transfer function,

\begin{equation}
\phi\left(\sum_jw_jt_j\right)=-\tau_m\log\left[\frac{\sum_je^{\frac{-(t_j+w_j)}{\tau_m}}+e^{\frac{-(-t_j-w_j)}{\tau_m}}}{\sum_je^{\frac{-(t_j-w_j)}{\tau_m}}+e^{\frac{-(-t_j+w_j)}{\tau_m}}}\right]
\label{eq:22}
\end{equation}

Converting $t_j$ and $w_j$ into differential domain as $w^+_j=a+w_j, w^-_j=a-w_j$ and $t^+_j=a+t_j, t^-_j=a-t_j$, which enables canceling of inherent device noise, we get

\begin{equation}
\phi\left(\sum_jw_jt_j\right)=-\tau_m\log\left[\frac{\sum_je^{\frac{-(t_j^++w_j^+)}{\tau_m}}+e^{\frac{-(t_j^-+w_j^-)}{\tau_m}}}{\sum_je^{\frac{-(t_j^++w_j^-)}{\tau_m}}+e^{\frac{-(t_j^-+w_j^+)}{\tau_m}}}\right]
\label{eq:23}
\end{equation}

The numerator and denominator can be written as,

\begin{eqnarray}
\sum_je^{\frac{-(t_j^++w_j^+)}{\tau_m}}+e^{\frac{-(t_j^-+w_j^-)}{\tau_m}} = e^{-\frac{t^+}{\tau_m}} \nonumber \\
\sum_je^{\frac{-(t_j^++w_j^-)}{\tau_m}}+e^{\frac{-(t_j^-+w_j^+)}{\tau_m}} = e^{-\frac{t^-}{\tau_m}}
\label{eq:24}
\end{eqnarray}

Substituting the approximation $f_a$, we get,

\begin{align}
\sum_j\vert{t^+-(t_j^++w_j^+)}\vert_++\vert{t^+-(t_j^-+w_j^-)}\vert_+=\tau_m \nonumber \\
\sum_j\vert{t^+-(t_j^++w_j^-)}\vert_++\vert{t^+-(t_j^-+w_j^+)}\vert_+=\tau_m
\label{eq:25}
\end{align}

Hence, the output $t$ of the TEMP neuron implementing the transfer function given in Eq. \ref{eq:11} is

\begin{equation}
t=\left\{ 
  \begin{array}{ c l }
    t^+-t^- & \textrm{if } t^+ \geq t^- \\
    0   &             \textrm{otherwise}
  \end{array}
\right.
\label{eq:26}
\end{equation}

\section{Appendix II}
\label{sec:appendix2}

\paragraph{Neuronal Dynamic of TEMP}
\begin{figure*}
\centering
\begin{subfigure}{0.49\textwidth}
\centering
\includegraphics[width=0.3\textwidth]{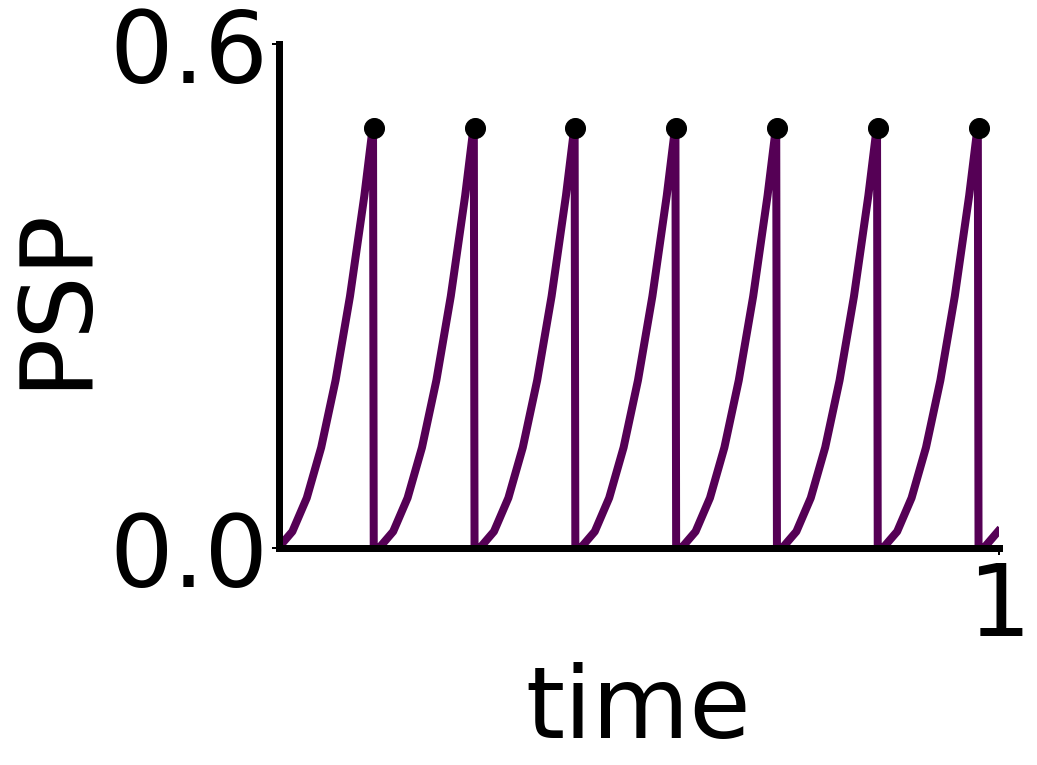}
\includegraphics[width=0.3\textwidth]{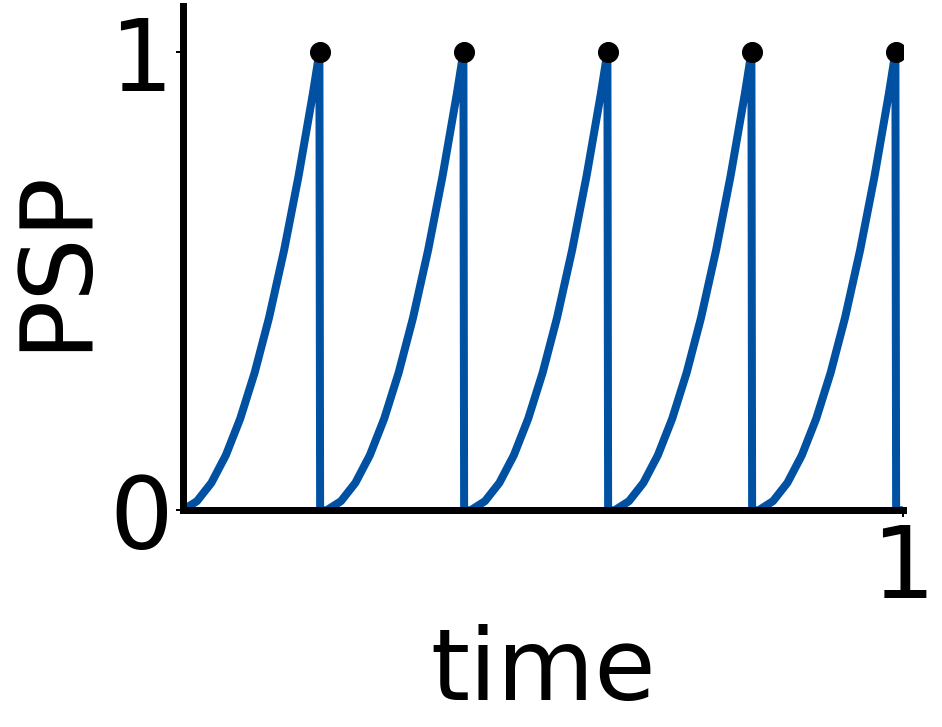}
\includegraphics[width=0.3\textwidth]{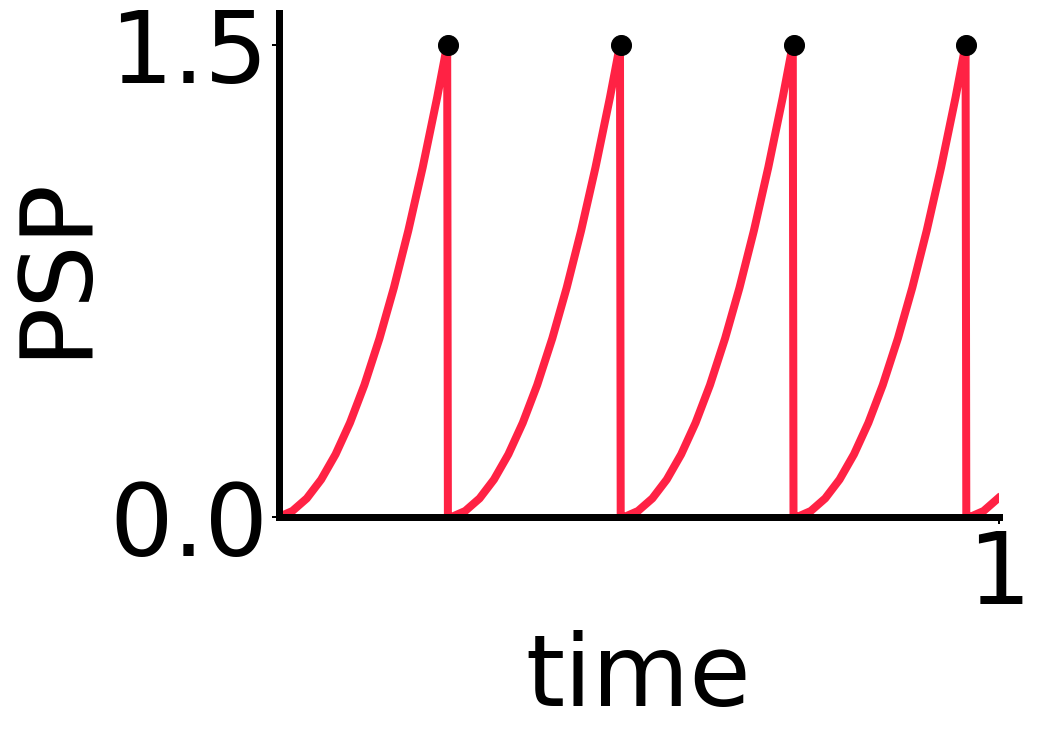}
\includegraphics[width=0.3\textwidth]{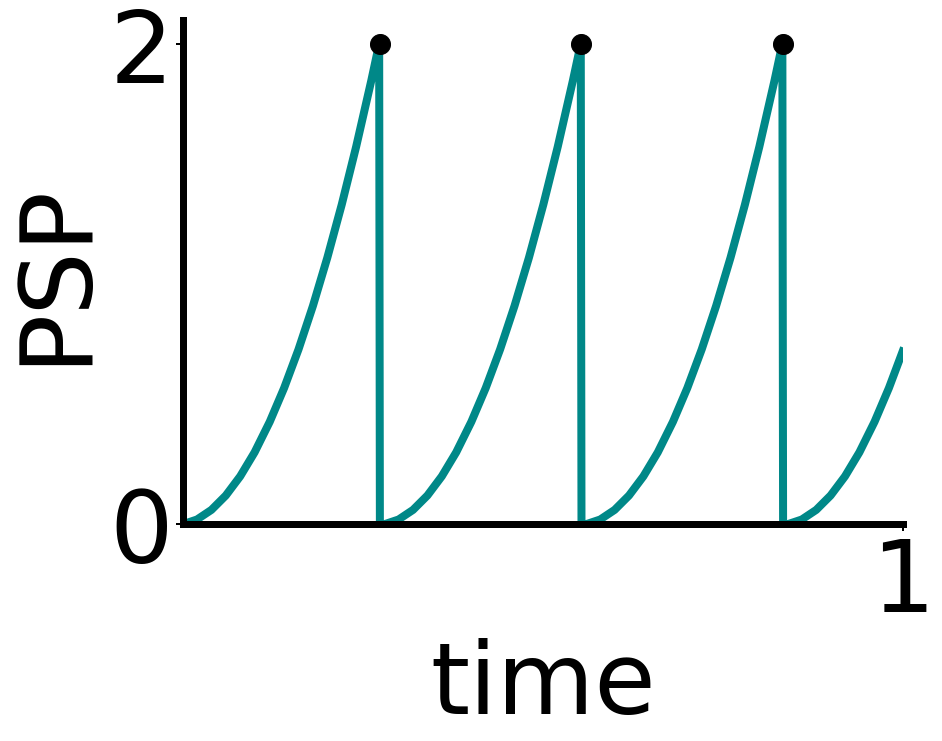}
\includegraphics[width=0.3\textwidth]{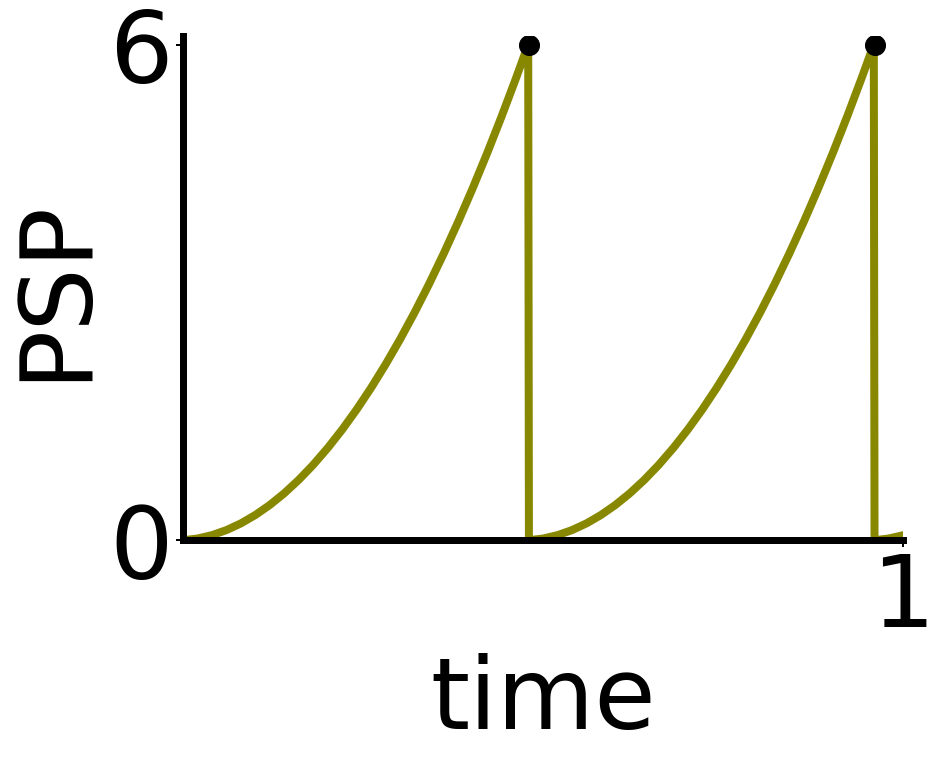}
\includegraphics[width=0.3\textwidth]{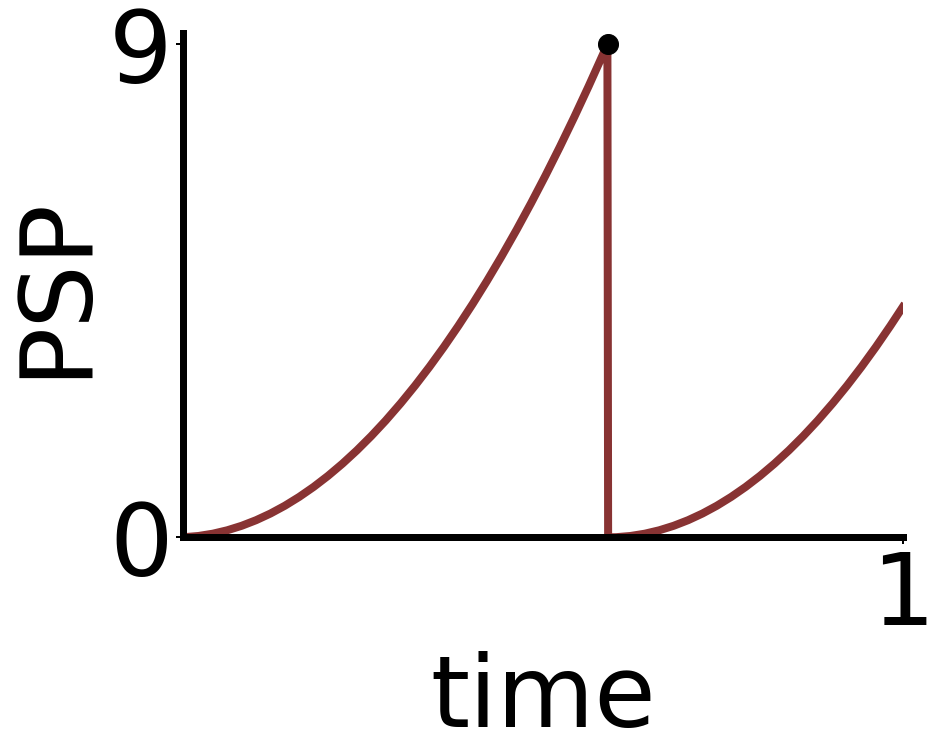}
\newline
\text{\small a)}
\end{subfigure}
\begin{subfigure}{0.49\textwidth}
\centering
\includegraphics[width=0.3\textwidth]{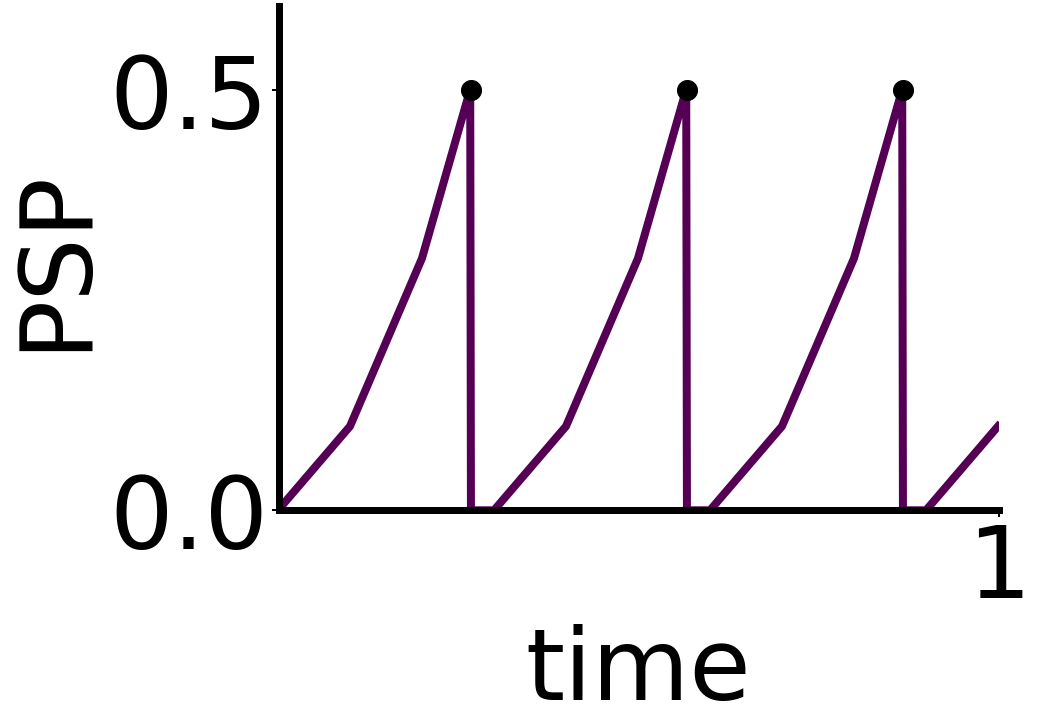}
\includegraphics[width=0.3\textwidth]{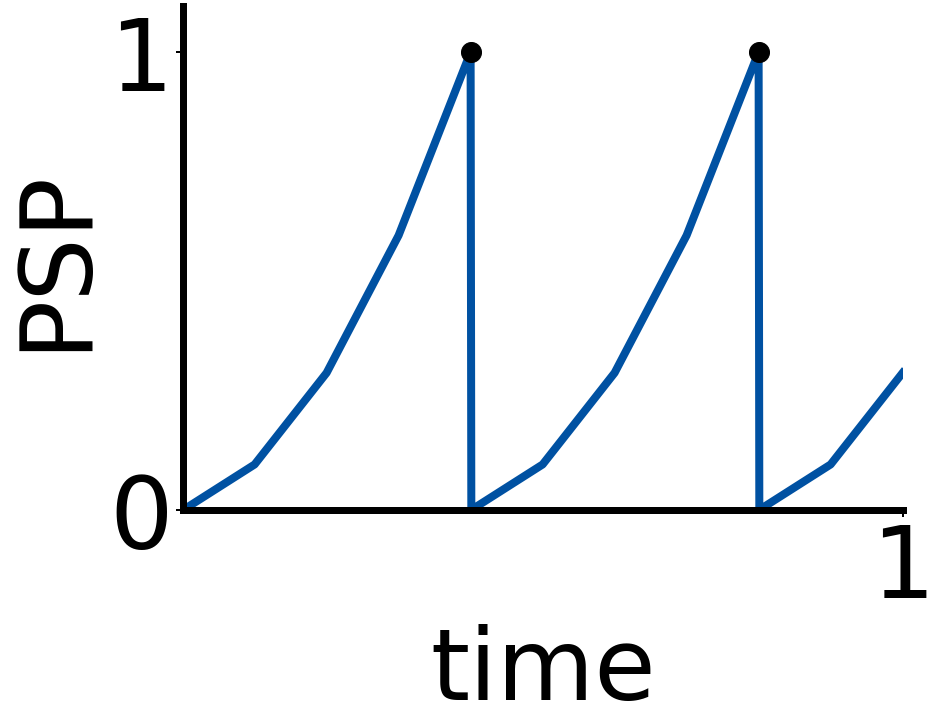}
\includegraphics[width=0.3\textwidth]{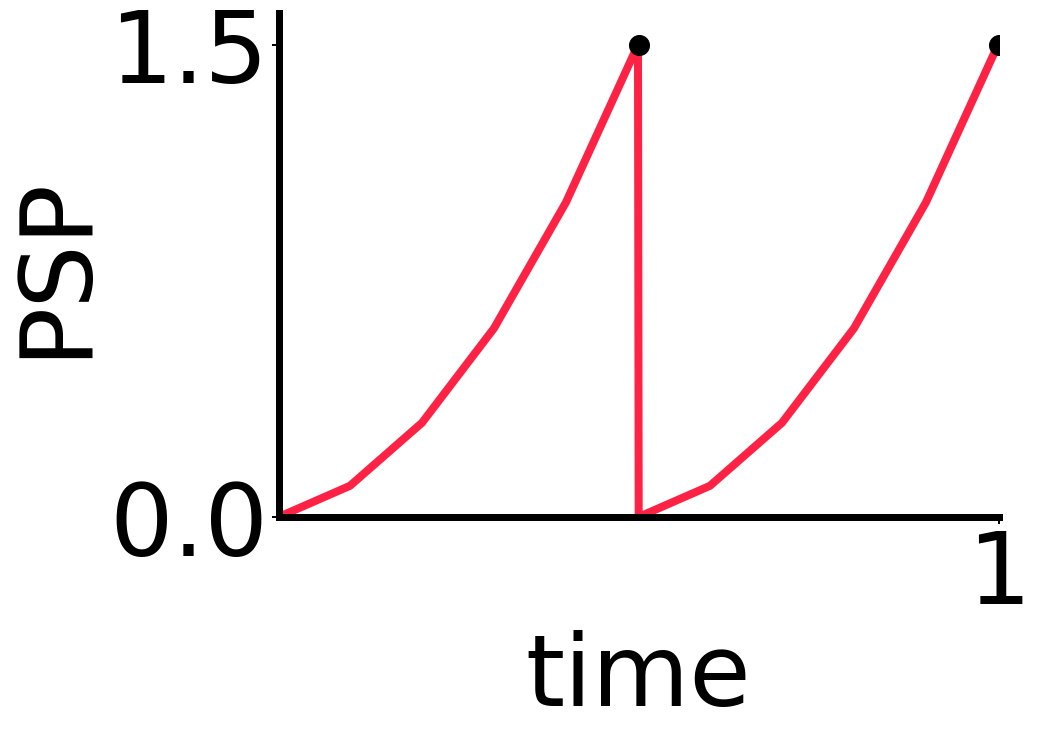}
\includegraphics[width=0.3\textwidth]{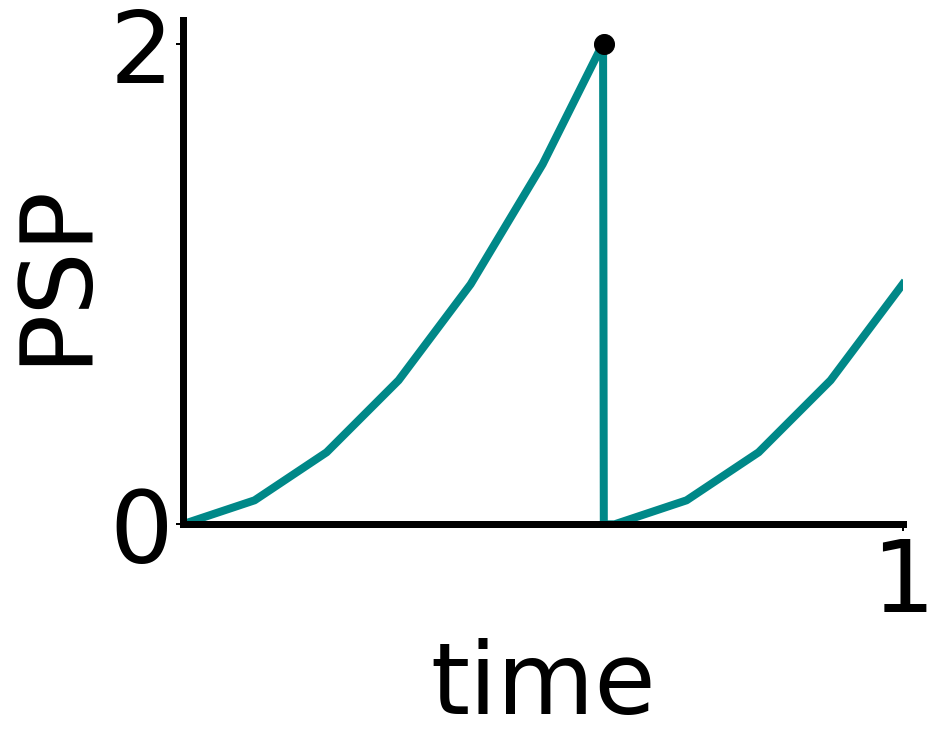}
\includegraphics[width=0.3\textwidth]{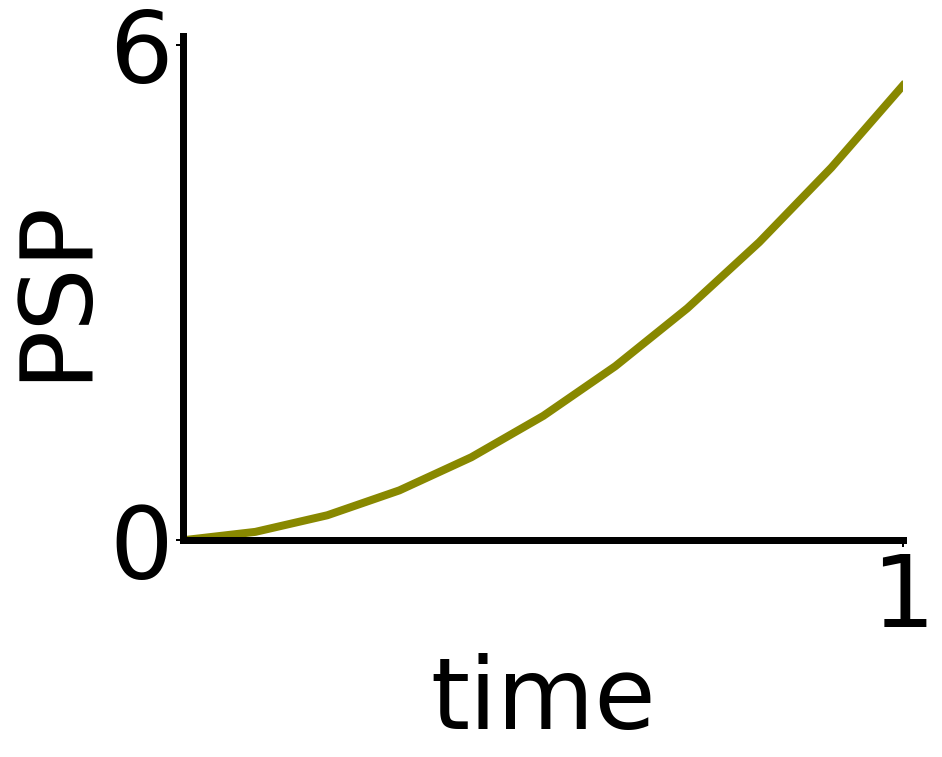}
\includegraphics[width=0.3\textwidth]{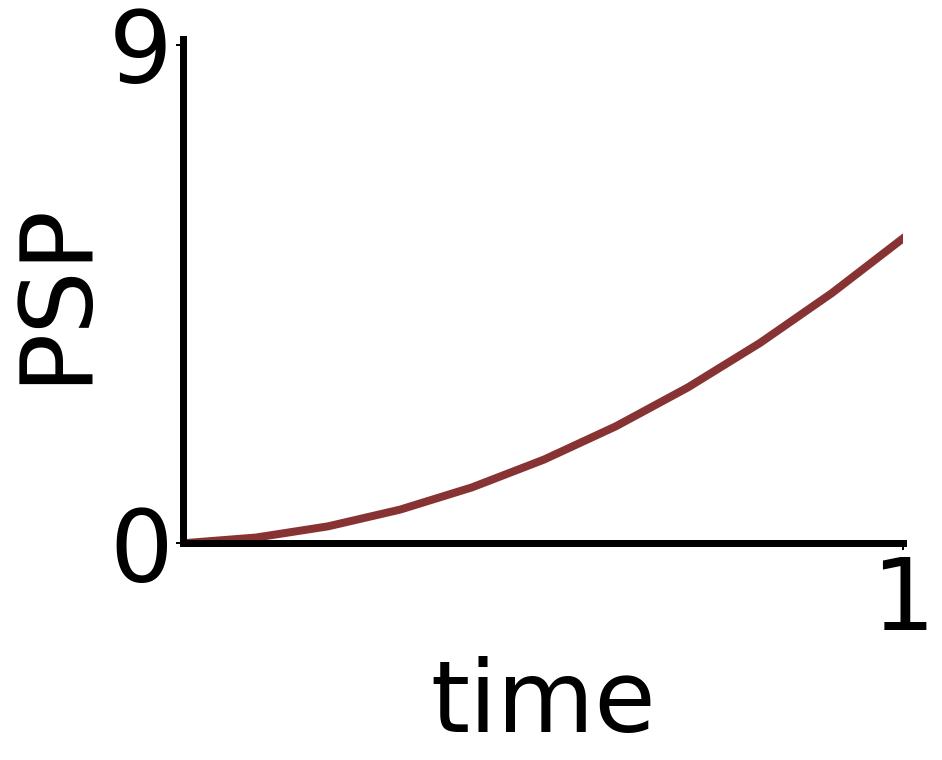}
\newline
\text{\small b)}
\end{subfigure}
\begin{subfigure}{0.49\textwidth}
\centering
\includegraphics[width=0.3\textwidth]{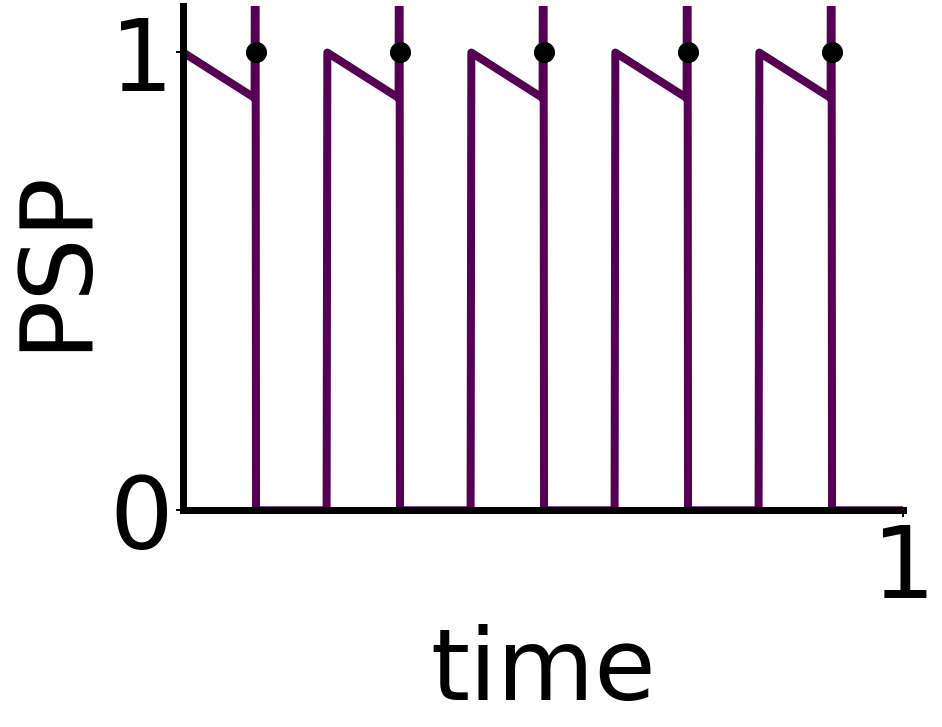}
\includegraphics[width=0.3\textwidth]{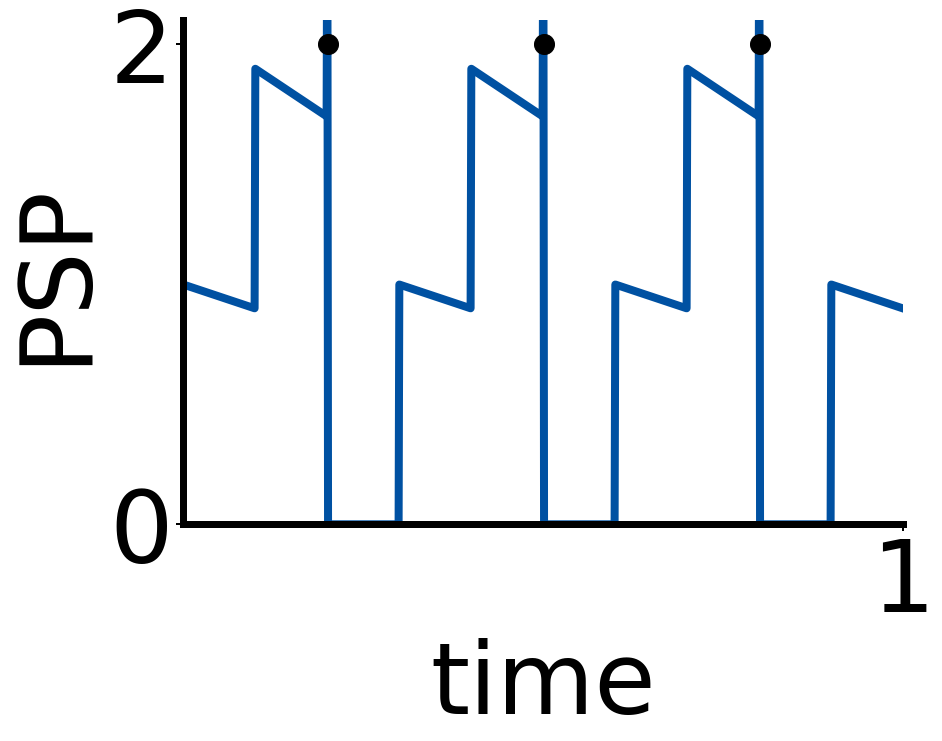}
\includegraphics[width=0.3\textwidth]{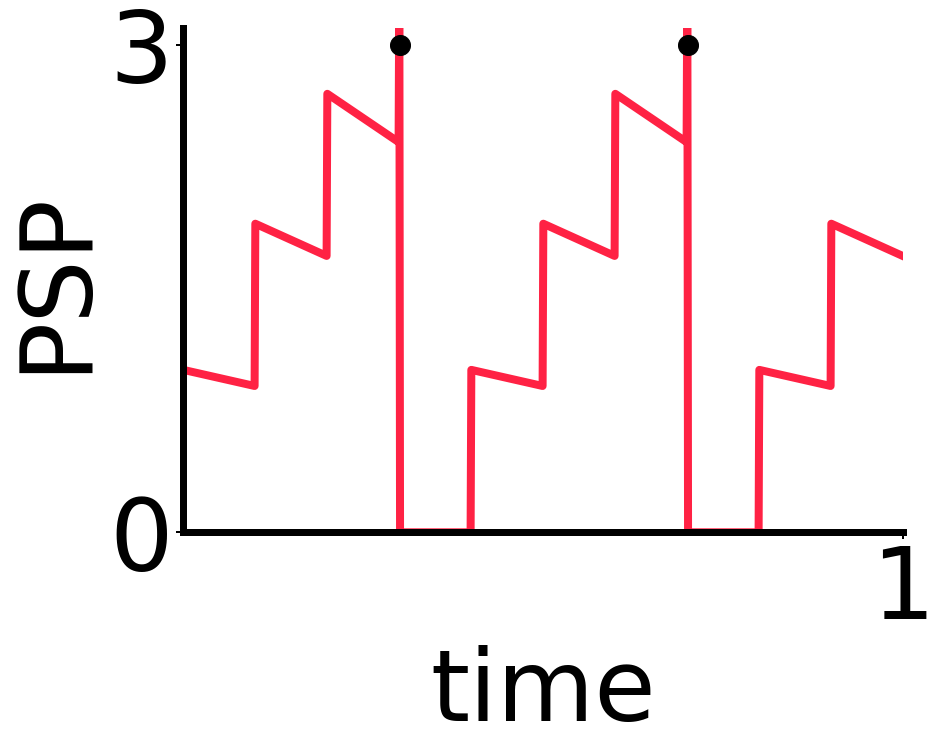}
\includegraphics[width=0.3\textwidth]{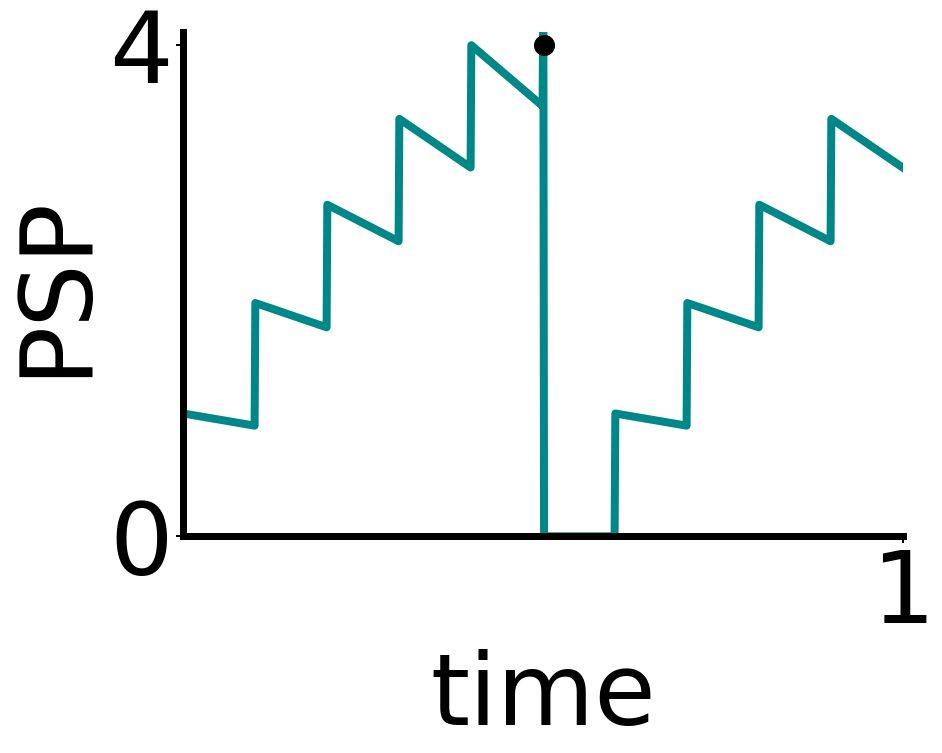}
\includegraphics[width=0.3\textwidth]{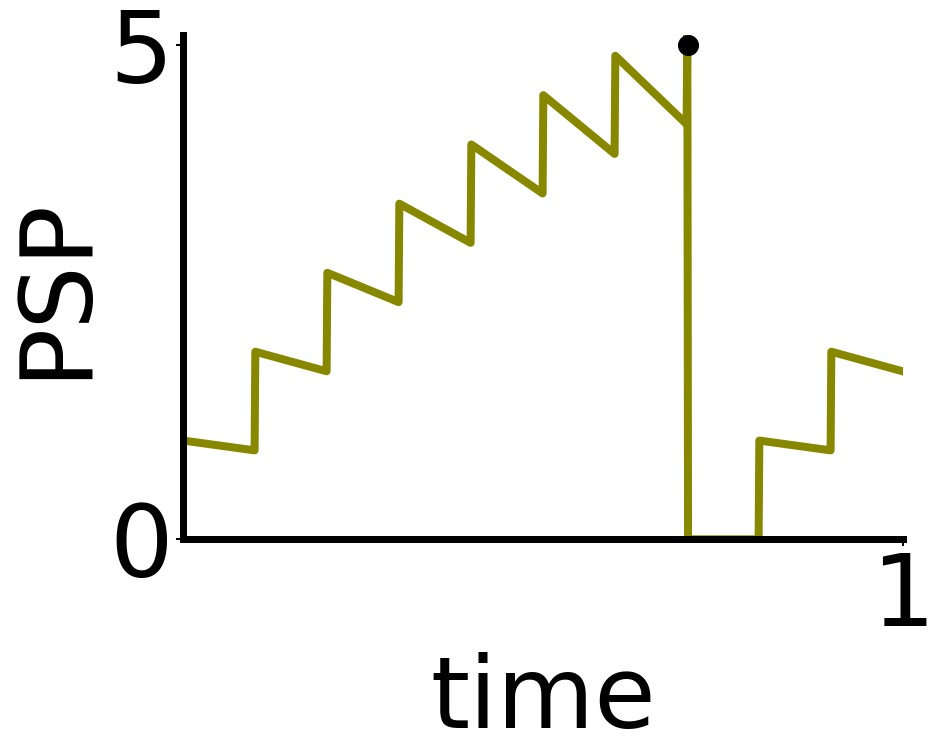}
\includegraphics[width=0.3\textwidth]{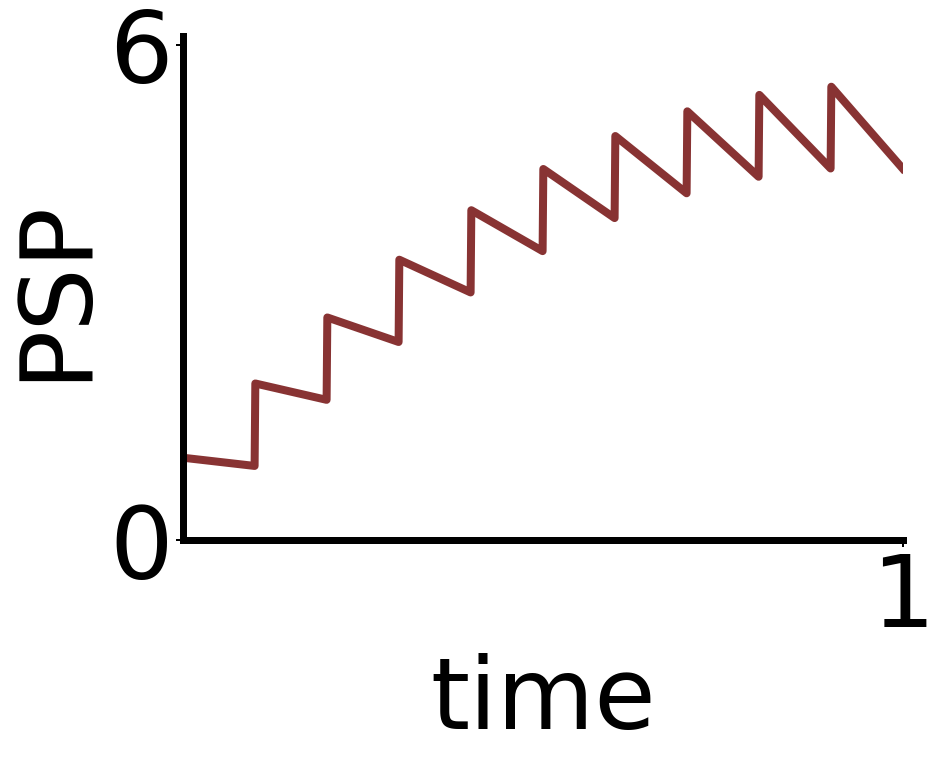}
\newline
\text{\small c)}
\end{subfigure}
\begin{subfigure}{0.49\textwidth}
\centering
\includegraphics[width=0.3\textwidth]{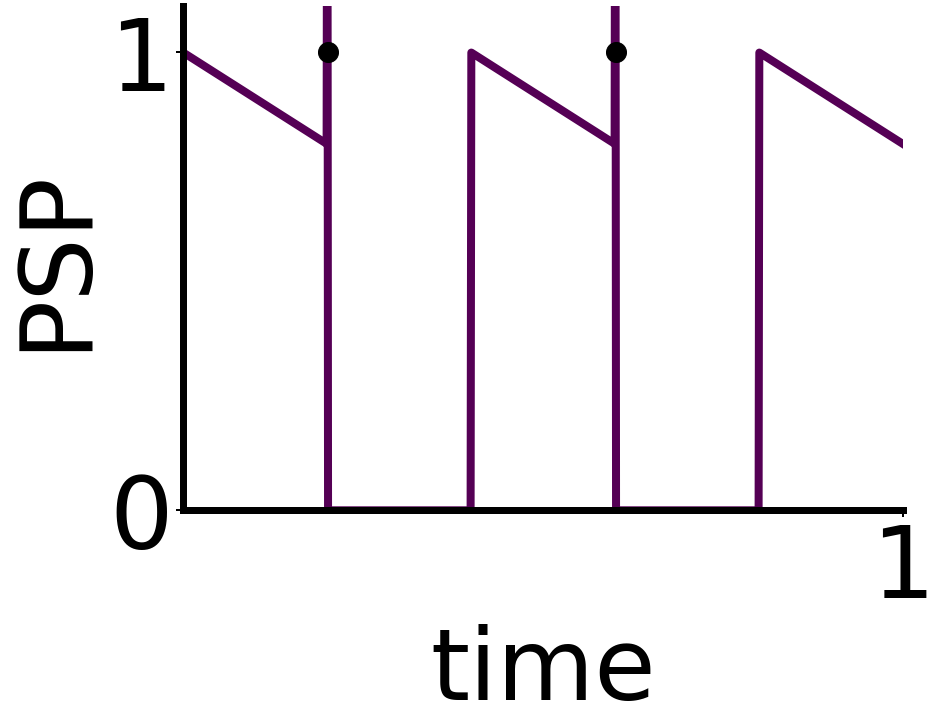}
\includegraphics[width=0.3\textwidth]{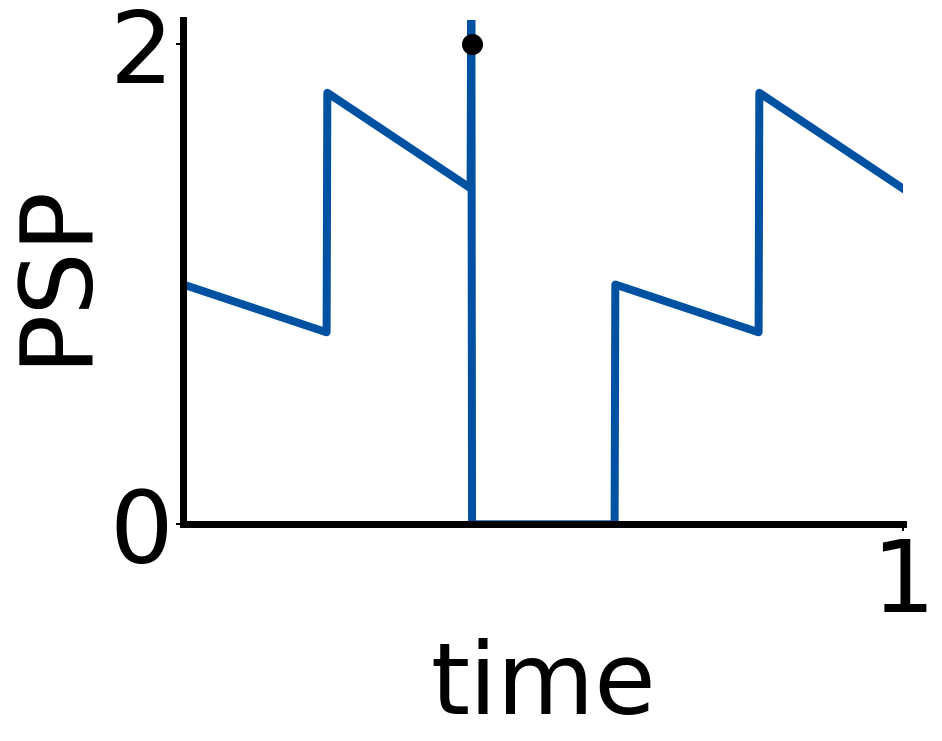}
\includegraphics[width=0.3\textwidth]{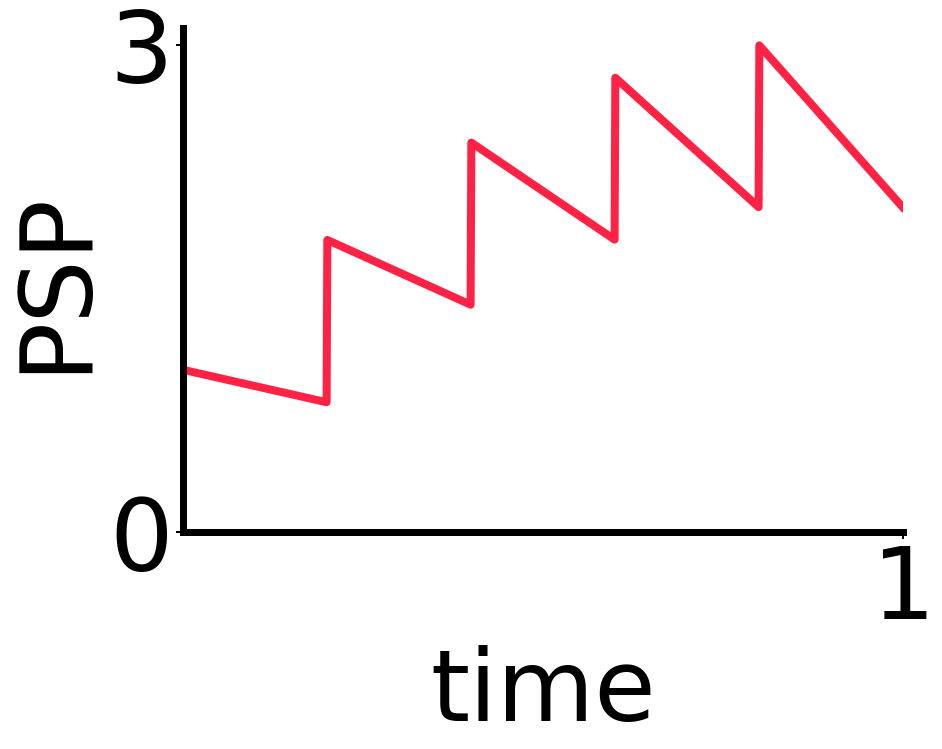}
\includegraphics[width=0.3\textwidth]{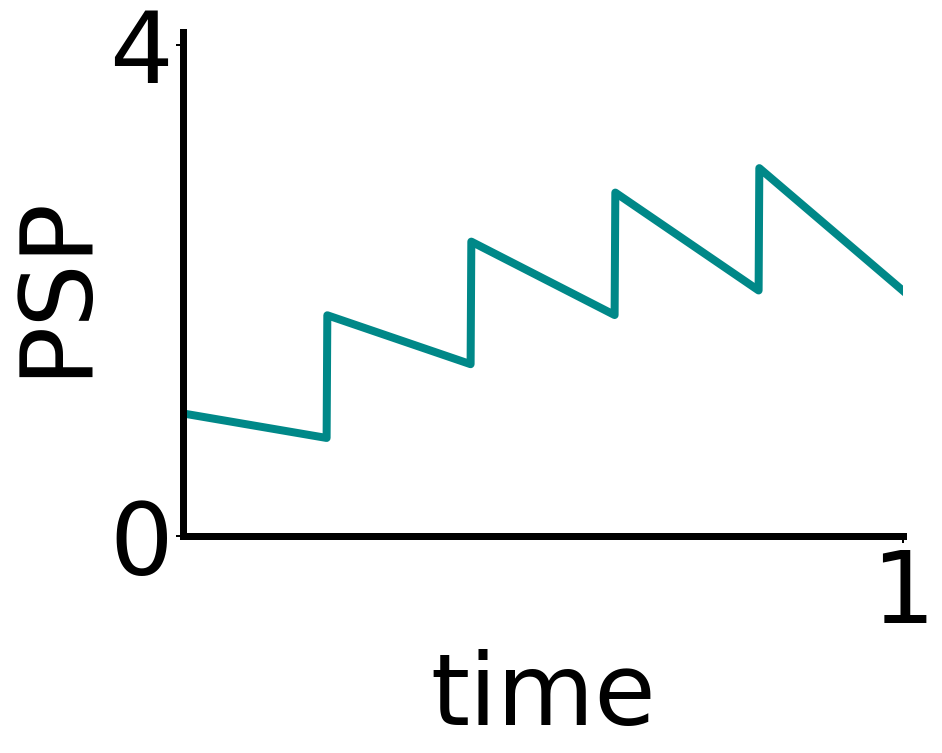}
\includegraphics[width=0.3\textwidth]{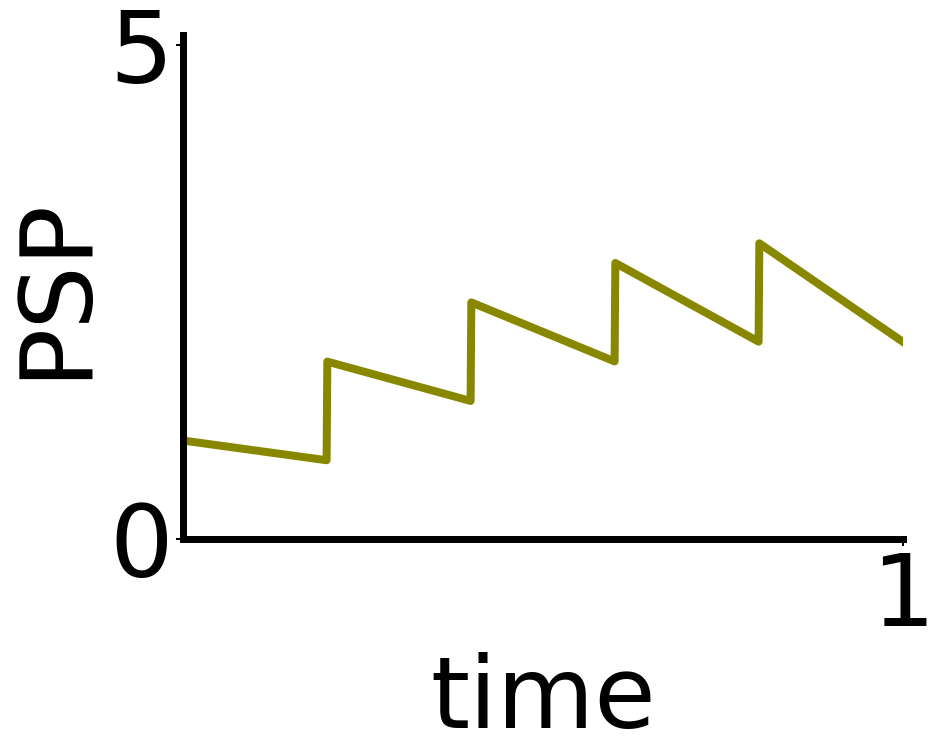}
\includegraphics[width=0.3\textwidth]{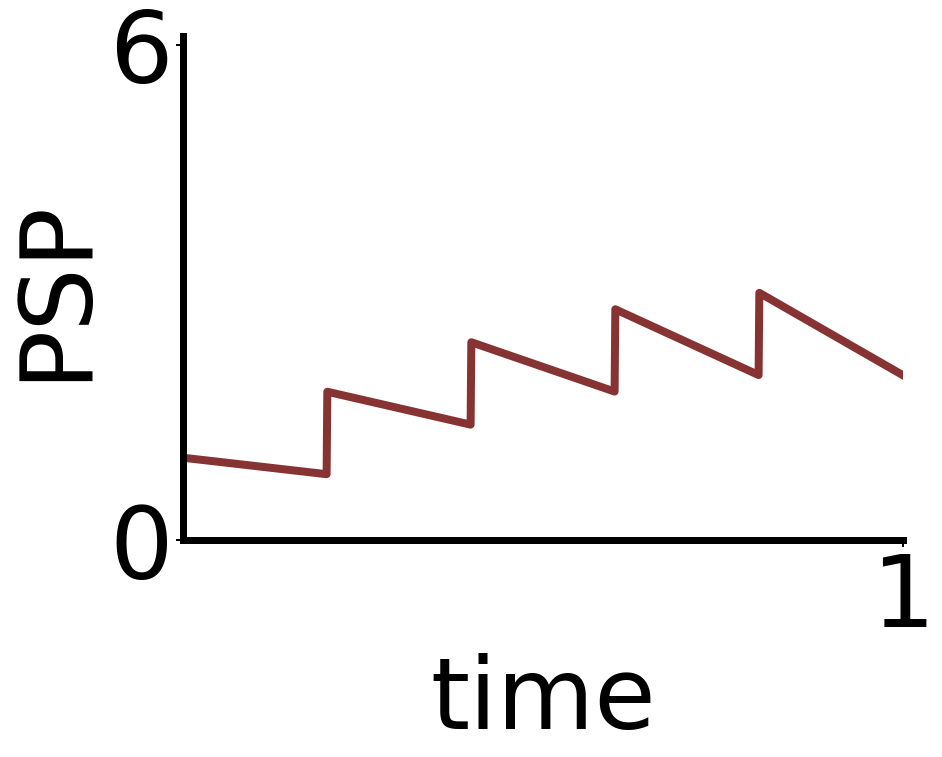}
\newline
\text{\small d)}
\end{subfigure}
\caption{\textbf{Neuronal dynamics of the proposed TEMP} in non-leaky (a,b) and leaky (c,d) mode for a pre-synaptic spike train generated at a frequency of $50$ Hz (a), $10$ Hz (b,c) and $5$ Hz (d). When the membrane potential reaches the threshold $\gamma$, an output spike is generated. The sub-plots show the variability of the output spike rate (output spike generation is marked as black dots) with respect to $\gamma$ and input spike rate.}
\label{fig:NeuralDynamics}
\end{figure*}
Note that TEMP can exhibit both leaky ($c-\sum_j\vert{t-t_i}\vert_+$) as well as non-leaky ($c+\sum_j\vert{t-t_i}\vert_+$) behavior. In the non-leaky model, input is retained until the neuron spikes, unlike the leaky version, where the potential starts to decay after reaching its peak. 

Fig. \ref{fig:NeuralDynamics} demonstrates the trajectories of the neural dynamics of TEMP, where we have evaluated the dynamics of the membrane potential of TEMP to input spike train $\delta(t-t_i)$ of varying frequencies. For each frequency, $\gamma$ has been varied and demonstrated that the relationship between the former and membrane potential is responsible for different spiking patterns.   
\paragraph{Extension of Non-linear classification abilities of TEMP}
\subparagraph{\textbf{Exclusive OR (XOR)}} 
\begin{figure*}
\centering
\begin{subfigure}{0.24\textwidth}
\centering
\includegraphics[width=\textwidth]{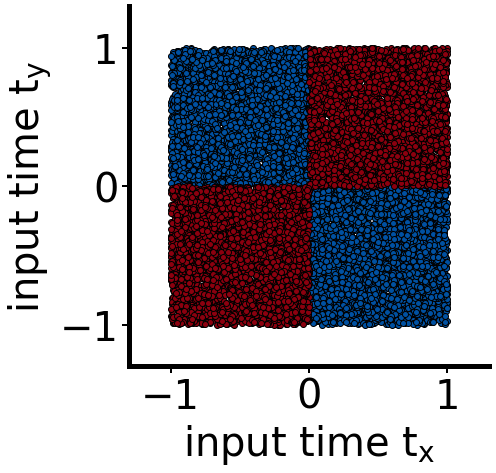}
\end{subfigure}
\begin{subfigure}{0.24\textwidth}
\includegraphics[width=\textwidth]{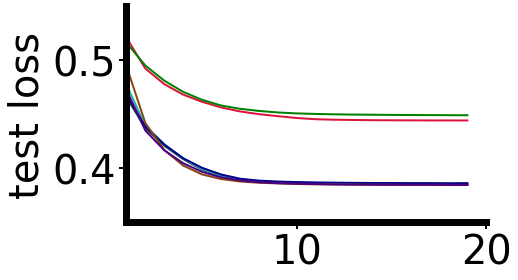}
\includegraphics[width=\textwidth]{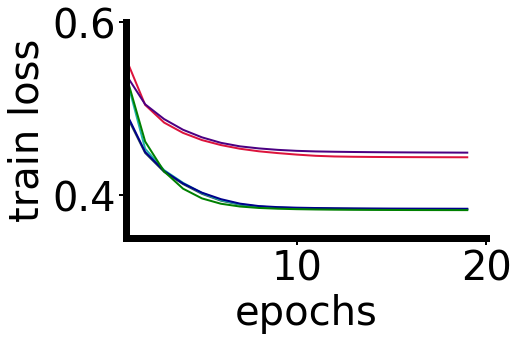}
\centering
\end{subfigure}
\begin{subfigure}{0.24\textwidth}
\centering
\includegraphics[width=\textwidth]{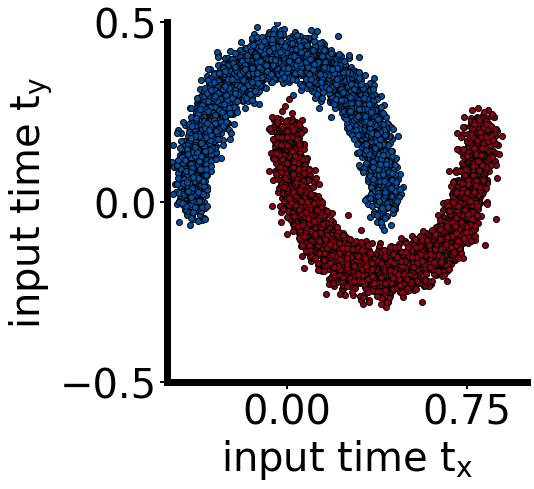}
\end{subfigure}
\begin{subfigure}{0.24\textwidth}
\centering
\includegraphics[width=\textwidth]{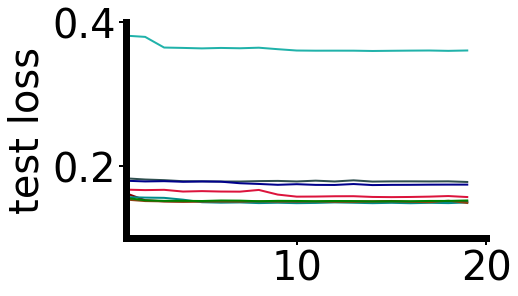}
\includegraphics[width=\textwidth]{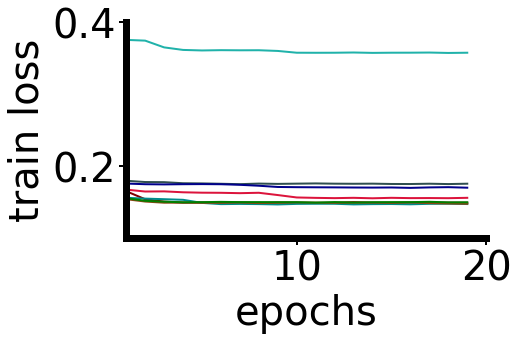}
\end{subfigure}
\begin{subfigure}{0.24\textwidth}
\centering
\includegraphics[width=\textwidth]{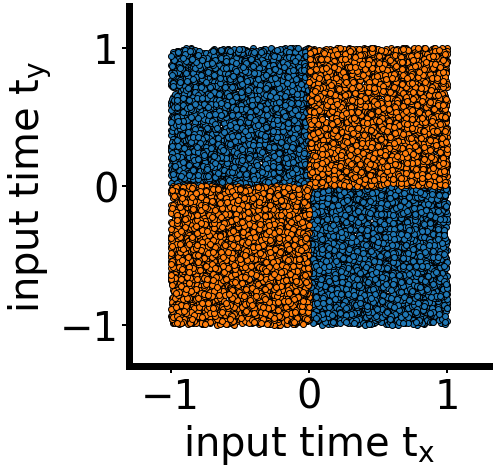}
\end{subfigure}
\begin{subfigure}{0.24\textwidth}
\includegraphics[width=\textwidth]{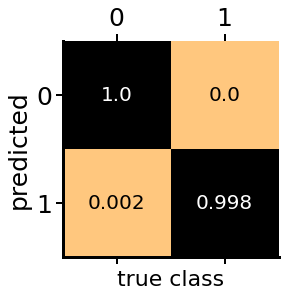}
\centering
\end{subfigure}
\begin{subfigure}{0.24\textwidth}
\centering
\includegraphics[width=\textwidth]{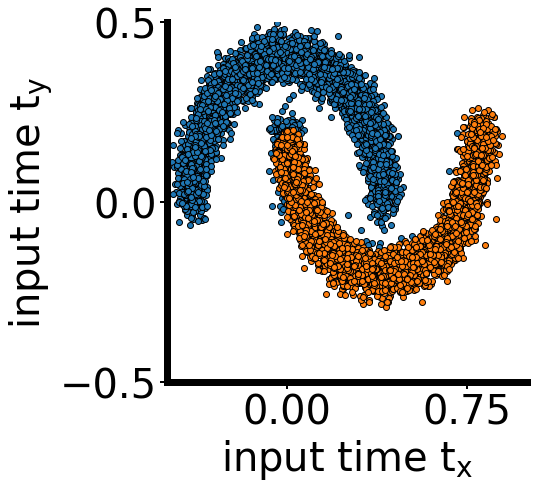}
\end{subfigure}
\begin{subfigure}{0.24\textwidth}
\centering
\includegraphics[width=\textwidth]{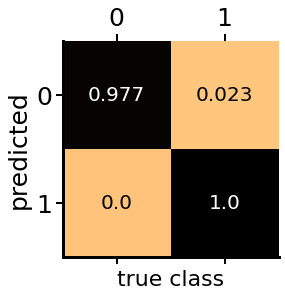}
\end{subfigure}

\begin{subfigure}{0.24\textwidth}
\centering
\includegraphics[width=\textwidth]{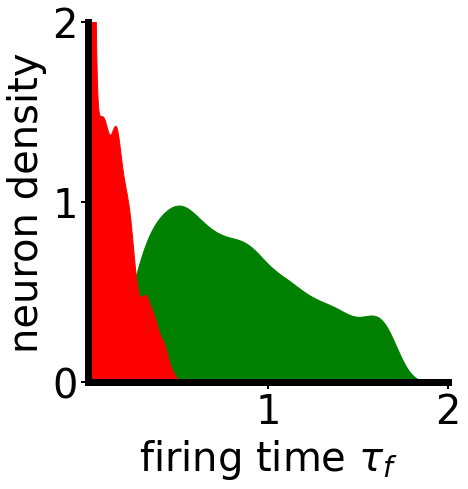}
\end{subfigure}
\begin{subfigure}{0.24\textwidth}
\centering
\includegraphics[width=\textwidth]{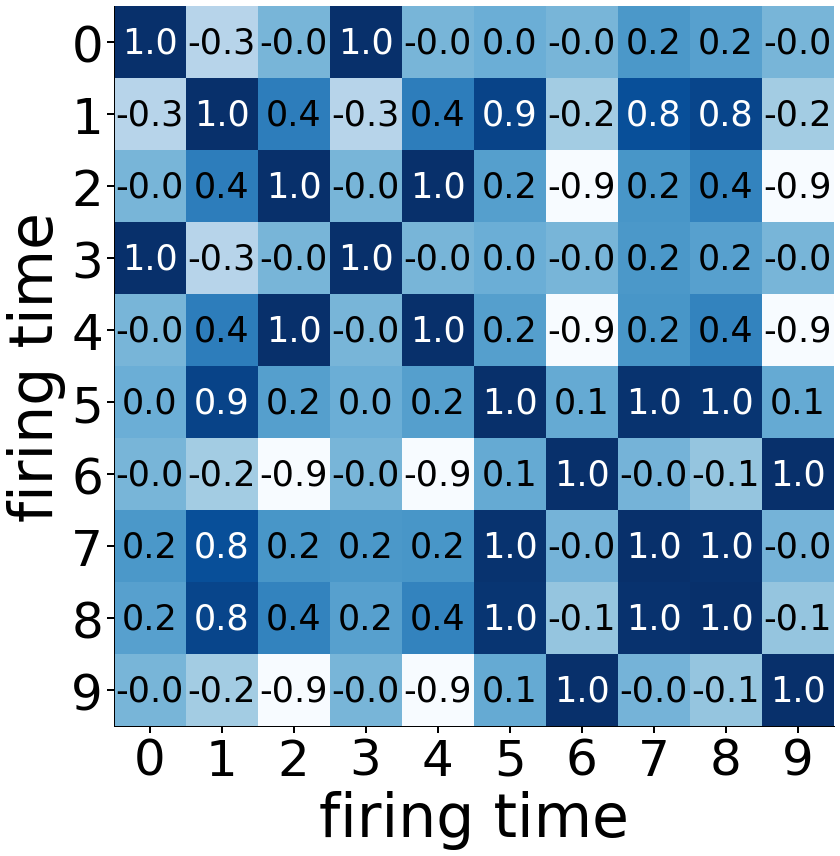}
\end{subfigure}
\begin{subfigure}{0.24\textwidth}
\centering
\includegraphics[width=\textwidth]{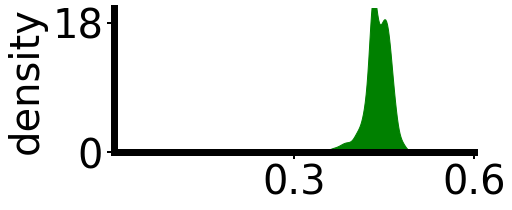}
\includegraphics[width=\textwidth]{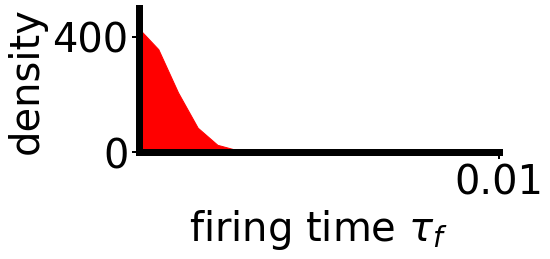}
\end{subfigure}
\begin{subfigure}{0.24\textwidth}
\centering
\includegraphics[width=\textwidth]{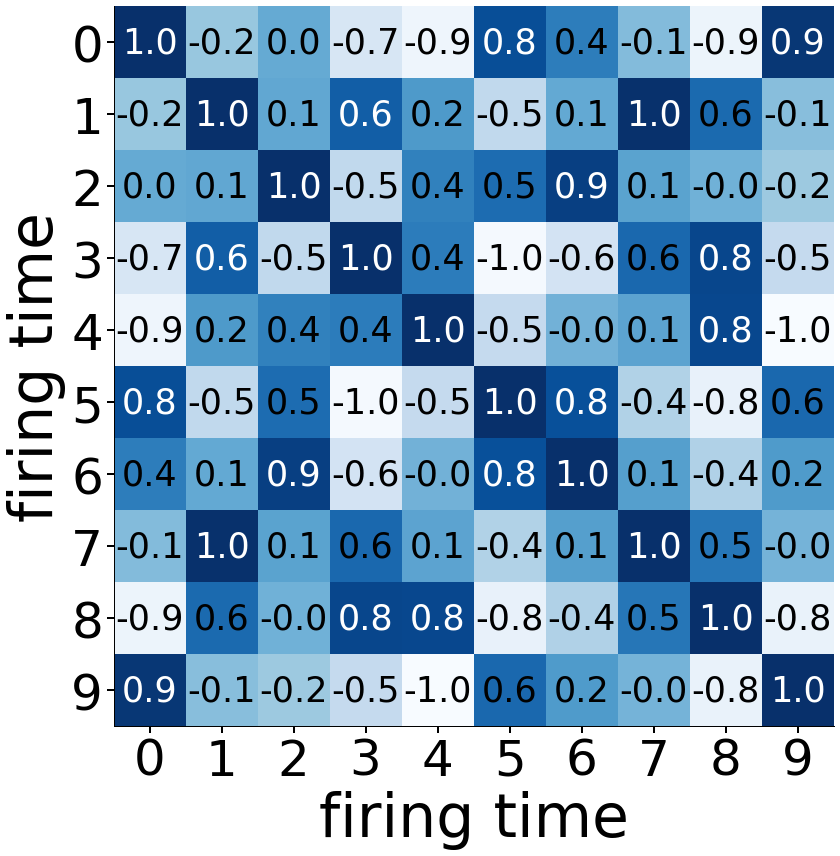}
\end{subfigure}
\caption{\textbf{Classification of XOR and MOON data set.} Top row: plots 1 and 3 - Illustration of XOR and MOON dataset. The input times $t_x$ and $t_y$ correspond to input spike train time corresponding to the $x$ and $y$ axes. Plot 2 and 4 - training progress in terms of loss on train and validation dataset for different runs (with different initializations) (XOR and MOON). Middle row: plots 1 and 3 - Classification result on XOR and MOON datasets. The color of each sample indicates the class predicted by the proposed solution. Plots 2 and 4 - confusion matrix of XOR and MOON. Bottom row: plots 1 and 3 - spike times of the two different class neurons. Our learning has induced a clear separation between the firing time of the two classes. Plots 2 and 4 - correlation between the firing time of $10$ hidden neurons of XOR and MOON dataset estimated across test samples. Very less value in non-diagonal elements indicates nearly zero correlation between the firing time of hidden neurons.}
\label{fig:MXclassify}
\end{figure*}

We begin by validating the proposed TEMP with a classic linearly non-separable XOR task. This has been done to verify the nonlinear classification capability of the proposed solution. XOR data has been generated from the uniform distribution $\mathcal{U}(-1,1)$ and encoded as differential spikes.

The architecture we have set up has a dense layer with ten TEMP and a classification layer with two TEMP, which is tasked to signal the true class. 

The network is initialized with values drawn from a normal distribution. Learning loss is defined such that TEMP belonging to true class fires $t^{-}$ far earlier than $t^{+}$. Loss is binary cross entropy loss between softmax activation of the classification layer and true class labels.

After training $20$ epochs with $60000$ samples (batch size of $128$) with Adam optimizer and an initial learning rate of $0.001$, the network successfully learned the XOR classification task with an accuracy of $99.6 \%$. The good classification accuracy proves that the non-linearity induced by the TEMP enables the classification of non-linearly separable XOR data (The results are presented in  Fig. \ref{fig:MXclassify}).
\begin{figure*}
\centering
\begin{subfigure}{\textwidth}
\centering
\includegraphics[width=0.19\textwidth]{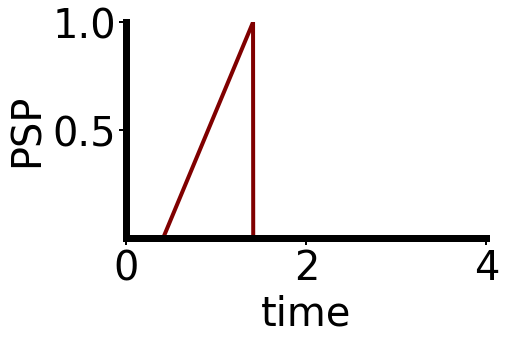}
\includegraphics[width=0.19\textwidth]{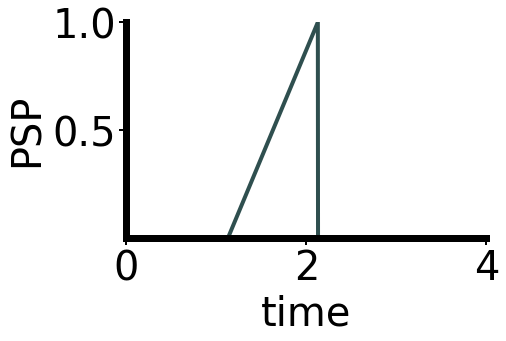}
\includegraphics[width=0.19\textwidth]{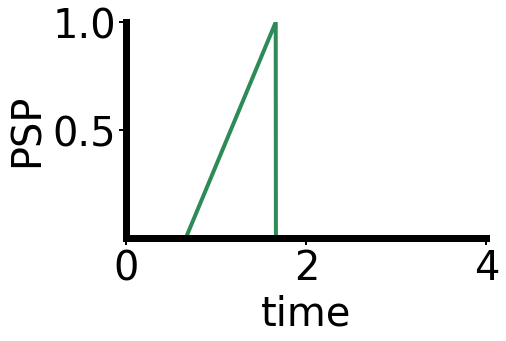}
\includegraphics[width=0.19\textwidth]{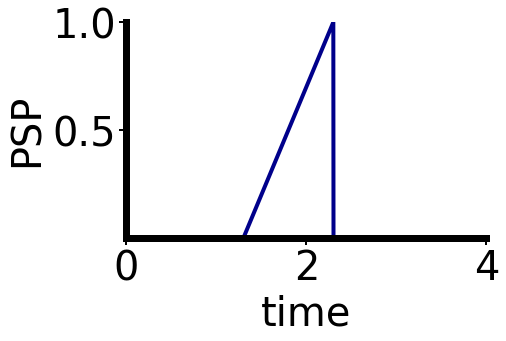}
\includegraphics[width=0.19\textwidth]{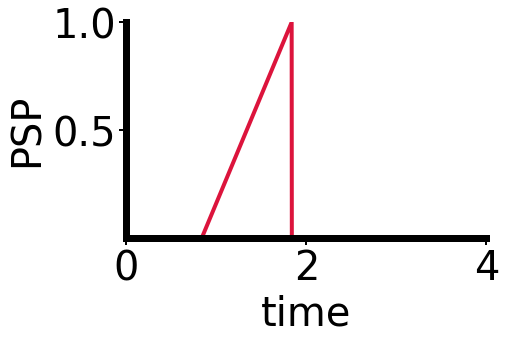}
\newline
\text{\small a)}\end{subfigure}
\begin{subfigure}{\textwidth}
\centering
\includegraphics[width=0.19\textwidth]{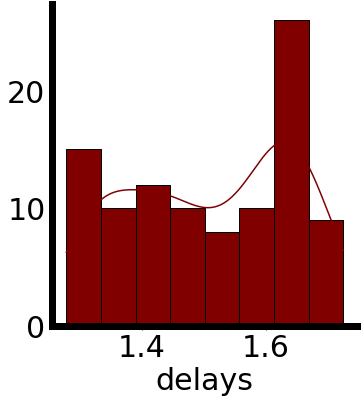}
\includegraphics[width=0.19\textwidth]{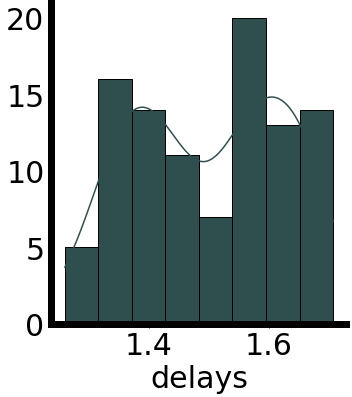}
\includegraphics[width=0.19\textwidth]{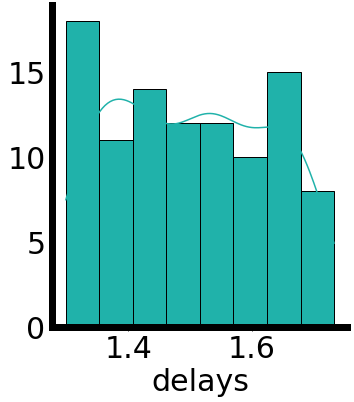}
\includegraphics[width=0.19\textwidth]{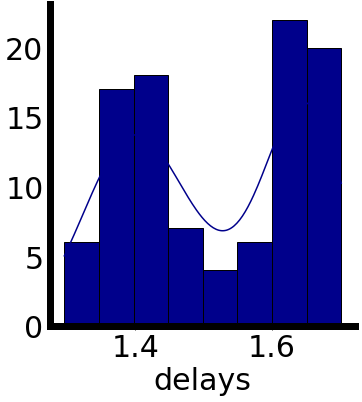}
\includegraphics[width=0.19\textwidth]{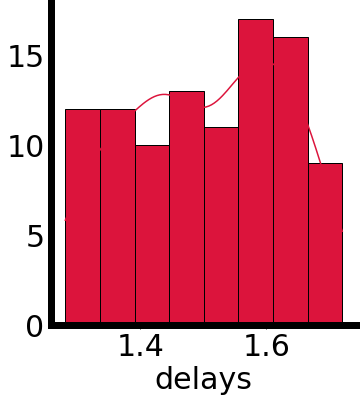}
\includegraphics[width=0.19\textwidth]{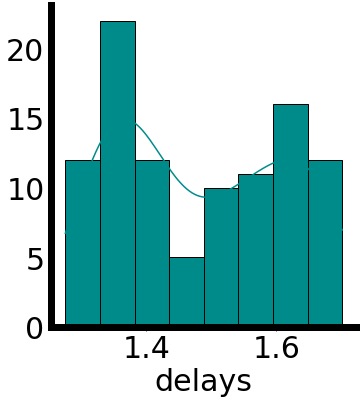}
\includegraphics[width=0.19\textwidth]{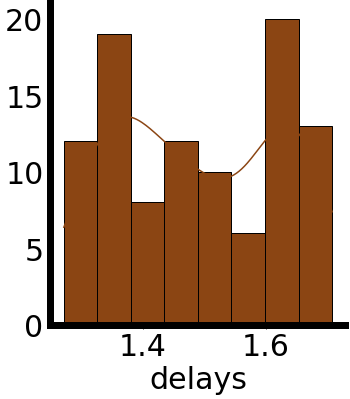}
\includegraphics[width=0.19\textwidth]{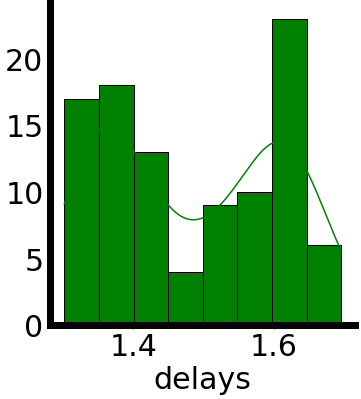}
\includegraphics[width=0.19\textwidth]{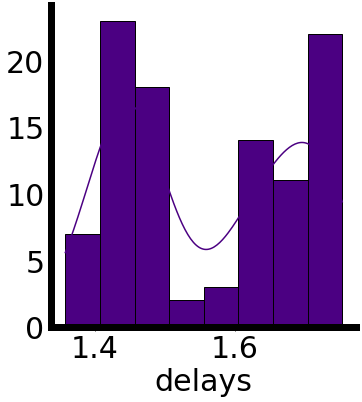}
\includegraphics[width=0.19\textwidth]{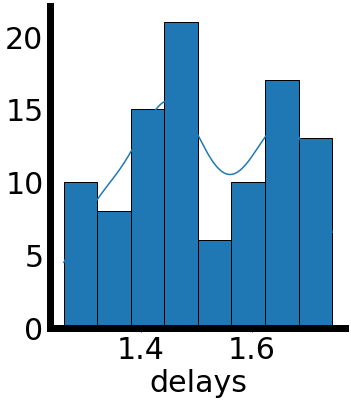}
\newline
\text{\small b)}
\end{subfigure}
\caption{\textbf{Benefits of the proposed TEMP.} (a): shows the membrane potential of 6 different stimuli from the XOR dataset. Before learning, these stimuli required two input spikes to trigger the post-synaptic neuron, whereas learning has reduced the number of spikes to one to reach the threshold. This indicates achieving the desired results with minimal spike computation. (b) the histogram plot of the delay (after learning) between hidden neurons to the $10$ class nodes of the network trained on the MNIST dataset. The difference in the delay distributions depicts the class-specific learning of delays.}
\label{fig:others}
\end{figure*}
\subparagraph{\textbf{MOON classification task}}

To further understand the effect of adding layers to classify non-linearly separable data, we have investigated the classification performance of the proposed TEMP solution with a synthetic two-dimensional binary classification dataset popularly known as the MOON dataset. We have implemented the following spiking network for this purpose: $2\rightarrow10\rightarrow20\rightarrow2$. The weights were initialized by drawing from the normal distribution. 

Training samples of $20000$ were presented to the network as mini-batches of size $128$ for $20$ epochs. The optimizer was set to Adam with a learning rate of $0.03$. It was necessary to include standardization ($\frac{x-\mu}{\sigma}$) as part of the loss function to sustain the propagation of gradients across the network during training. The proposed TEMP demonstrated success in fitting the data with a test accuracy of $99.25\pm{0.03}\%$, thereby emphasizing the capability of the proposed TEMP to estimate optimal nonlinear boundary on test data (The results are presented in  Fig. \ref{fig:MXclassify}).

\paragraph{Additional experiments for sparsity in causal spikes}

Fig. \ref{fig:others}a shows the dynamics of TEMP for a sample set of XOR data input spike train. With all the other conditions remaining the same, the input spike train was delayed with random and learned delay. Randomly delayed scenarios mandated two spikes to trigger firing, whereas spikes delayed with learned $w$ could go into firing mode with the arrival of a single spike. Delays, when applied to input spikes, shift the effect of a less informative input spike to a later time (before which TEMP is triggered), thus reducing the number of required spike processing. The introduction of delay in TEMP helps to remove undesired input spikes from the output spike generation process.

\paragraph{Learning of class-specific delay}

Fig. \ref{fig:others}b gives insight into the distribution of delay learned by the synapses connecting $100$ hidden nodes and $10$ class nodes for MNIST data. It brings out the class-specific distribution of delay.

\section{Data Availability }
The code for implementing the TEMP inference engine is available at \url{https://github.com/NeuRonICS-Lab/temp-framework}.

\bibliography{sn-bibliography}% common bib file


%\newpage

\end{document}